\documentclass[letterpaper, 10pt, conference]{ieeeconf}
\IEEEoverridecommandlockouts
\usepackage[utf8]{inputenc}
\usepackage[english]{babel}

\usepackage{comment}
\usepackage{xspace}
\usepackage{graphicx}
\usepackage{overpic}
\usepackage{svg}
\usepackage[dvipsnames]{xcolor}
\usepackage{bm}
\usepackage[hang,flushmargin]{footmisc}
\usepackage{adjustbox}
\usepackage{etoolbox}
\usepackage{environ}
\usepackage{pbalance}
\usepackage{flushend}
\usepackage{mathtools}
\usepackage[normalem]{ulem}
\usepackage{hhline}
\usepackage{tabularx}
\usepackage{multirow}
\usepackage{booktabs}
\usepackage{colortbl}
\usepackage{float}
\usepackage{placeins}
\usepackage{array}
\usepackage{threeparttable}
\usepackage{makecell}

\usepackage{amsmath}
\usepackage{amssymb}
\usepackage{amsthm}
\usepackage{siunitx}
\usepackage{mathtools}

\usepackage{csquotes}
\usepackage[
	maxbibnames=4,
	maxcitenames=2,
	natbib=true,
	bibstyle=ieee,
	citestyle=numeric-comp,
	backend=biber,
	sorting=none,
	giveninits=true,
	url=false,
	doi=false,
	eprint=false,
	isbn=false,
]{biblatex}

\DeclareMathOperator*{\argmin}{arg\,min}

\makeatletter
\patchcmd{\@makecaption}{\scshape}{}{}{}
\newcommand{\onetagright}{\tagsleft@false}
\makeatother

\usepackage{subcaption}
\newtheoremstyle{main}
{1em}                                                %
{1em}                                                %
{\itshape}                                           %
{0pt}                                                %
{\scshape}                                           %
{\\*}                                                %
{2pt}                                                %
{\thmname{#1}\thmnumber{ #2}: \thmnote{\itshape #3}} %

\usepackage[linesnumbered,ruled,noend]{algorithm2e}
\newcommand{\removelatexerror}{\let\@latex@error\@gobble}

\let\labelindent\relax
\usepackage[inline]{enumitem}

\usepackage[
	activate   = {true},
	protrusion = false,
	expansion  = true,
	kerning    = true,
	spacing    = true,
	tracking   = false,
	auto       = true,
	selected   = true,
	factor     = 1000,
	stretch    = 10,
	shrink     = 10,
]{microtype}

\usepackage{csquotes}

\definecolor{purduegold}{HTML}{C28E0E} %

\makeatletter
\let\NAT@parse\undefined
\makeatother
\usepackage[pdfa,colorlinks,bookmarksopen,bookmarksnumbered,allcolors=purduegold]{hyperref}
\usepackage{bookmark}

\usepackage[nameinlink,capitalise]{cleveref}
\crefname{line}{line}{lines}
\crefname{figure}{Fig.}{Figs.}
\Crefname{figure}{Fig.}{Figs.}
\crefname{equation}{Eq.}{Eqs.}
\Crefname{equation}{Eq.}{Eqs.}
\crefname{section}{Sec.}{Secs.}
\Crefname{section}{Sec.}{Secs.}
\crefname{definition}{Def.}{Defs.}
\Crefname{definition}{Def.}{Defs.}
\crefname{algorithm}{Alg.}{Algs.}
\Crefname{algorithm}{Alg.}{Algs.}
\crefname{assumption}{Asm.}{Asms.}
\Crefname{assumption}{Asm.}{Asms.}
\crefname{subassumption}{Asm.}{Asms.}
\Crefname{subassumption}{Asm.}{Asms.}
\crefname{problem}{Problem}{Problems}
\Crefname{problem}{Problem}{Problems}

\makeatletter
\newcommand\footnoteref[1]{\protected@xdef\@thefnmark{\ref{#1}}\@footnotemark}
\makeatother

\graphicspath{ {./images/} }

\providecommand{\Opt}{\texttt{(Opt)}}

\providecommand{\best}[1]{\textbf{#1}}
\providecommand{\snd}[1]{\underline{#1}}

\title{\fontsize{17pt}{24pt}\selectfont \bf Stochastic Neural Signed Swept Volume for Real-time Chance-Constrained Trajectory Optimization}
\newif\ifanonymous
\anonymousfalse

\ifanonymous
  \author{Anonymous Author(s)}
  \hypersetup{hidelinks}
\else
    \author{
    Qingyi Chen, Kevin Zhang, Lucas Chen, and Zachary Kingston
    \thanks{QC, KZ, LC, and ZK are with the Department of Computer Science, Purdue University, West Lafayette, IN, USA. {\tt \{chen5221, zhan4196, chen4007, zkingston\}@purdue.edu}. 
    }}
\fi

\begin{document}
\maketitle

\begin{abstract}
  Collision-free motion planning requires reliable collision models from sensed environments and validation of states along a continuous trajectory.
  To make this tractable, most planners check for collision at discrete states along continuous trajectories against a single determinized model of the environment, introducing a trade-off between safety and computational efficiency.
  While continuous collision checking approaches that approximate the swept volume of the robot exist, they are computationally expensive or overly conservative.
  Data-driven approaches can learn the swept volume; however, these neural models are susceptible to approximation errors and are therefore often limited to serving as coarse filters for downstream collision checkers.
  In this work, we propose to learn a signed distance function of the swept volume as a probabilistic field, enabling quantification of epistemic uncertainty, incorporation of perception noise, and eventual integration into a chance-constrained trajectory optimization framework.
  We demonstrate our approach on challenging high-dimensional manipulation problems with significant sensor noise, both in simulation and on real hardware.
\end{abstract}

\setlength{\abovedisplayskip}{3pt} %
\setlength{\belowdisplayskip}{3pt} %
\setlength{\abovedisplayshortskip}{3pt} %
\setlength{\belowdisplayshortskip}{3pt} %

\setlength{\floatsep}{0pt}%
\setlength{\textfloatsep}{8pt}%
\setlength{\intextsep}{8pt} %

\setlength{\belowcaptionskip}{0.5em} %

\setlength{\dbltextfloatsep}{0pt} %
\setlength{\dblfloatsep}{0pt} %

\section{Introduction and Related Work}

Collision avoidance is a fundamental requirement for robot motion planning, especially in safety-critical and human-centric environments. To achieve this, a robot must efficiently compute continuous collision-free motions while accounting for uncertainty arising from imperfect sensing and dynamic obstacles. Conventional planning methods, such as sampling-~\cite{lavalle2001rapidly,thomason2024motions,kuffner2000rrt} and trajectory optimization-based methods~\cite{sundaralingam2023curobo,ratliff2009chomp} typically check collision over discretized trajectories for computational tractability.
While increasing discretization density improves safety, it is computationally expensive, particularly in high-dimensional configuration spaces where collision queries dominate planning time~\cite{bialkowski2011massively}. 
Furthermore, these methods do not explicitly account for uncertainty in sensed geometry or future obstacle motion, further complicating reliable collision avoidance in real-world settings.

Swept volume representations address the first challenge by describing the entire region occupied by a robot throughout a motion. However, exact swept volumes are expensive to compute and difficult to scale to articulated manipulators~\cite{peternell2005swept, larsen2000fast}.
Classical planning approaches therefore rely on conservative or local approximations. TrajOpt~\cite{schulman2014motion} uses convex hull-based over-approximations over short trajectory segments, while reachability-based methods~\cite{holmes2020reachable,michaux2024safe} bound swept occupancy through set propagation, and other approaches employ related geometric constructions~\cite{ashur2026spite}. These methods improve upon point-wise collision checking but remain computationally expensive and limited to short trajectories.

\begin{figure}[t]
    \centering
    \begin{subfigure}[t]{0.31\columnwidth}
        \centering
        \includegraphics[width=\linewidth, trim=21 27 3 36, clip]{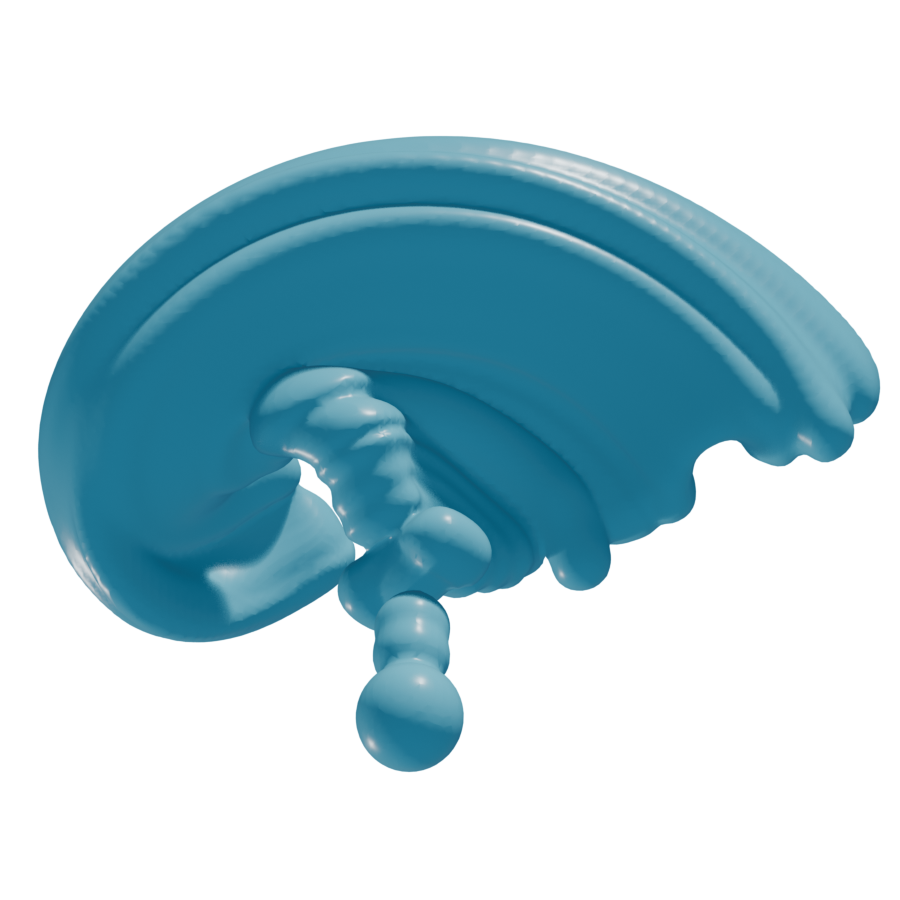}\\[3pt]
        \includegraphics[width=\linewidth, trim=21 27 3 36, clip]{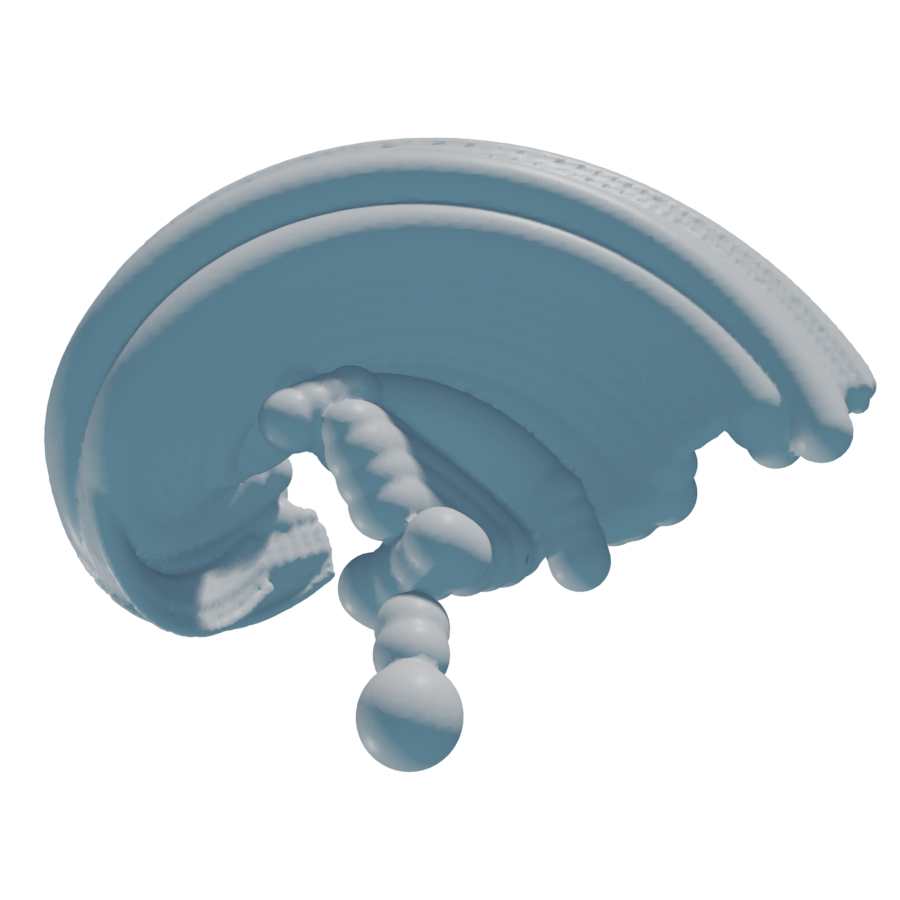}\\[3pt]
        \includegraphics[width=\linewidth, trim=21 27 3 36, clip]{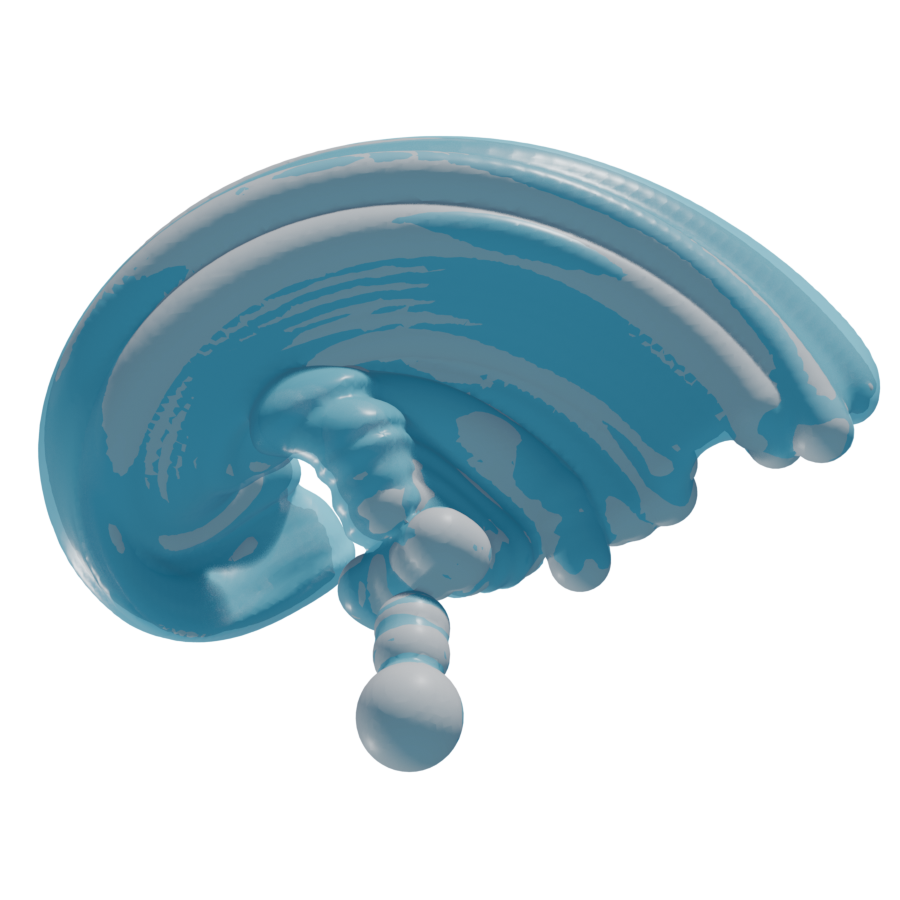}
        \caption{Franka Panda}
        \label{fig:reconstruction_panda}
    \end{subfigure}
    \hspace{0.03\columnwidth}
    \begin{subfigure}[t]{0.22\columnwidth}
        \centering
        \includegraphics[width=\linewidth, trim=91 36 189 32, clip]{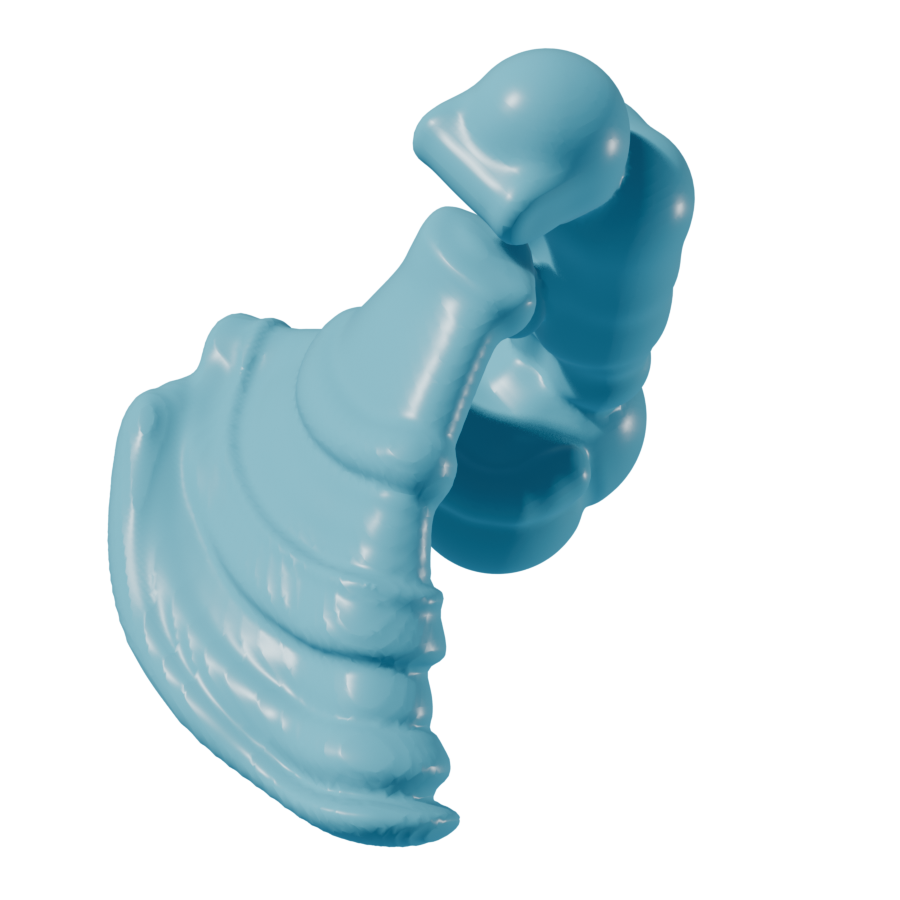}\\[3pt]
        \includegraphics[width=\linewidth, trim=91 36 189 32, clip]{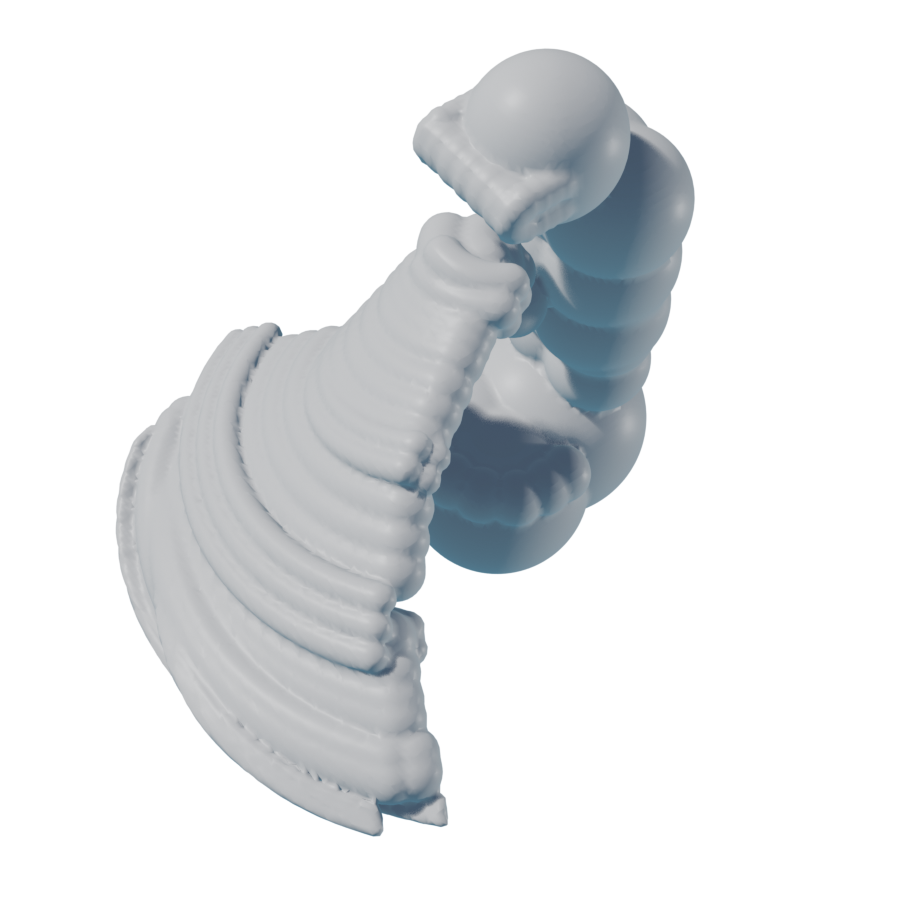}\\[3pt]
        \includegraphics[width=\linewidth, trim=91 36 189 32, clip]{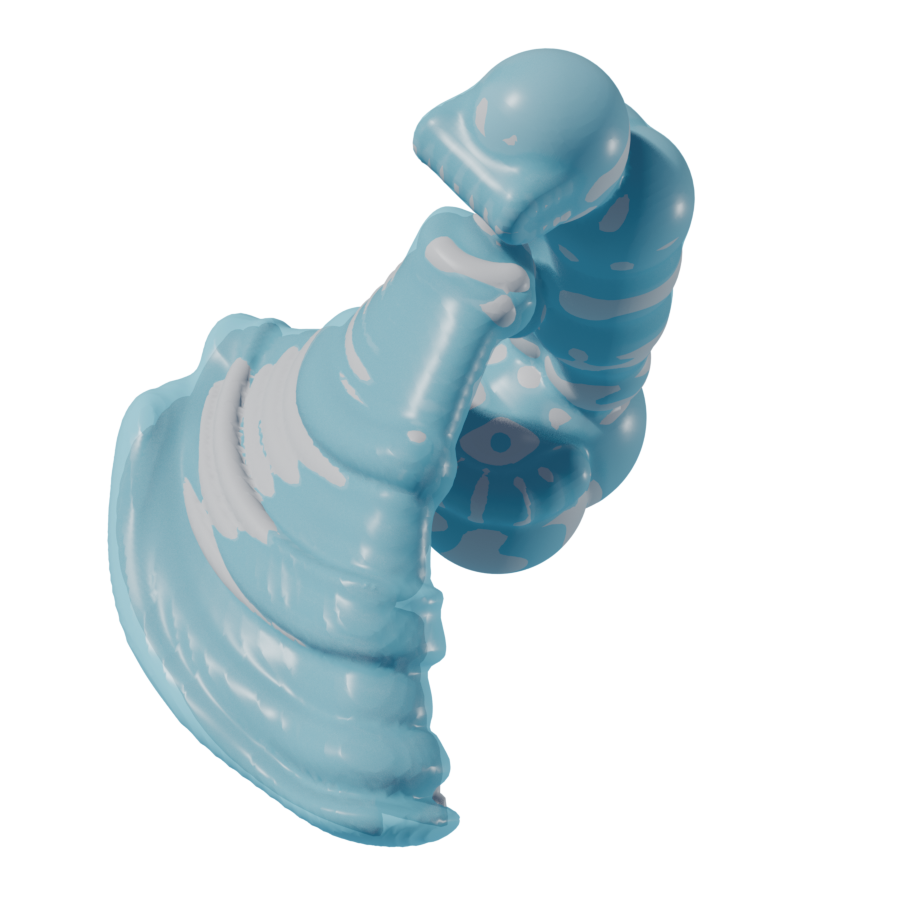}
        \caption{Fetch}
        \label{fig:reconstruction_fetch}
    \end{subfigure}
    \hspace{0.03\columnwidth}
    \begin{subfigure}[t]{0.26\columnwidth}
        \centering
        \includegraphics[width=\linewidth, trim=122 59 69 36, clip]{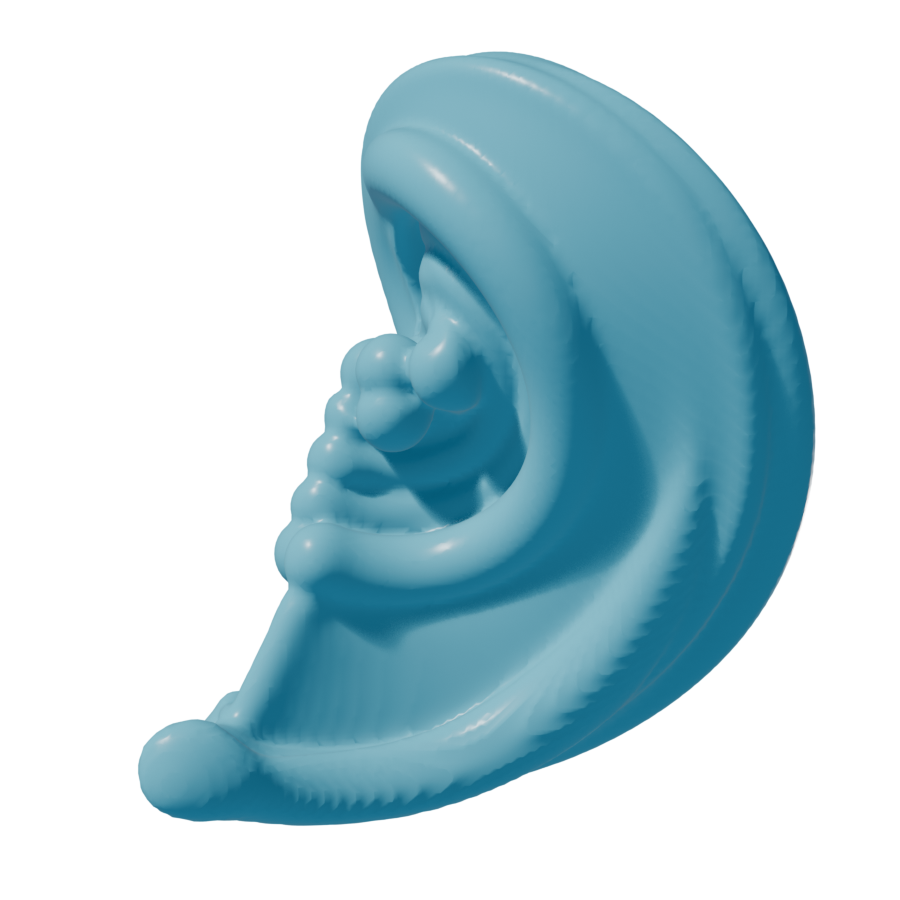}\\[3pt]
        \includegraphics[width=\linewidth, trim=122 59 69 36, clip]{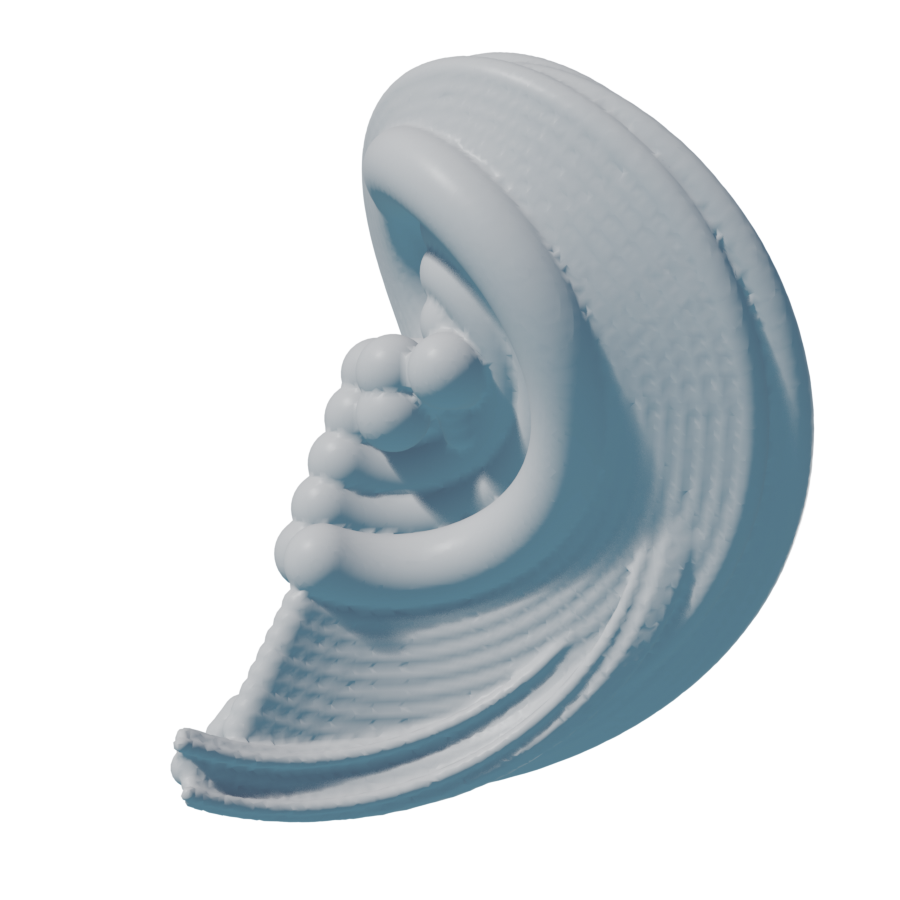}\\[3pt]
        \includegraphics[width=\linewidth, trim=122 59 69 36, clip]{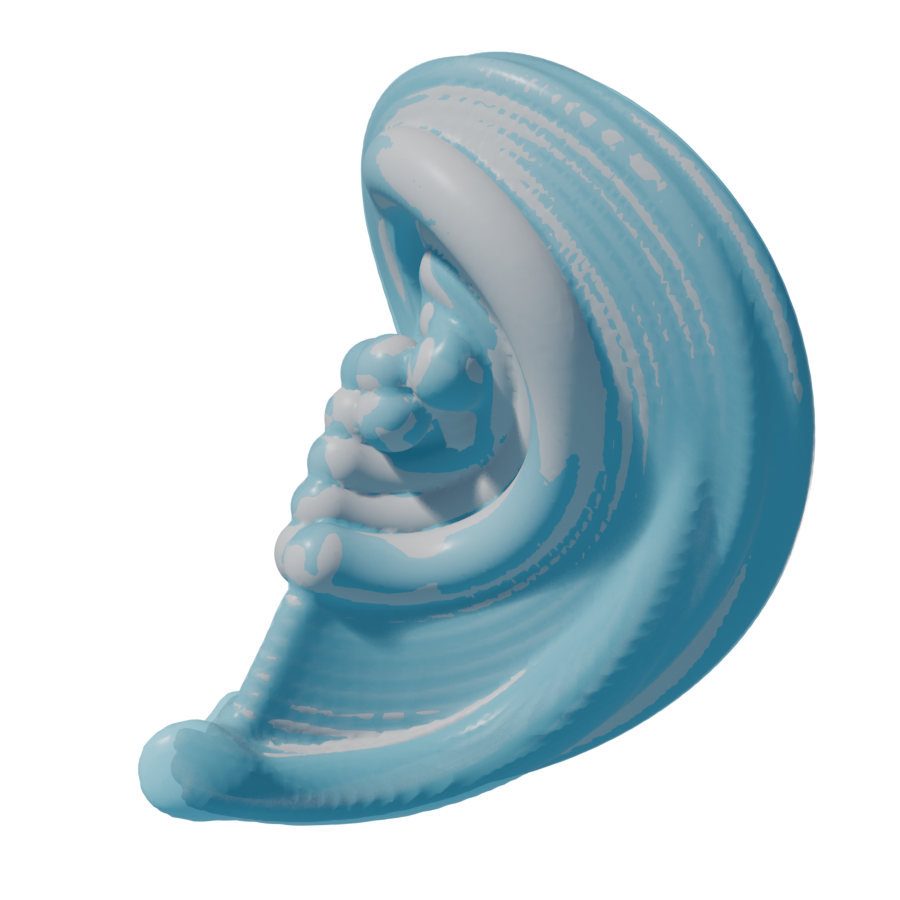}
        \caption{UR5}
        \label{fig:reconstruction_ur5}
    \end{subfigure}
    \vspace{-1em}
    \caption{
    Our method uses a learned swept volume model to formulate continuous collision-avoidance constraints for uncertainty-aware motion planning. The figure shows reconstructed neural swept volumes for the three robots. Within each group, the columns show the predicted swept volume (blue), the ground-truth swept volume (white), and an overlay visualization.}
    \label{fig:reconstruction}
\end{figure}

Recent learning-based approaches propose neural approximations of swept volumes in light of their efficiency and scalability~\cite{fast-deep-sv-estimator, joho2024neural, pmlr-v305-jung25e}, typically representing a signed distance function learned from geometric mesh unions~\cite{joho2024neural} or  reachability computation~\cite{michaux2023reachability}.
However, deterministic neural approximations remain susceptible to modeling error and cannot account for perception uncertainty. As a result, they cannot be directly trusted within planning pipelines without additional safety mechanisms. Prior work often relies on conservative post-hoc calibration~\cite{kwon2025conformalized} or uses learned models only as coarse filters preceding exact geometric collision checking~\cite{joho2024neural}, but these strategies reduce the computational advantages of the learned representation and do not provide a unified mechanism for incorporating both model and perception uncertainty.

Uncertainty-aware motion planning has separately been studied for uncertain dynamics, perception, and environment estimation~\cite{brudermuller2026cc,quintero2024stochastic,liu2023radius,luders2010chance,meng2022nr}. Chance-constrained and stochastic planning methods formulate collision avoidance probabilistically rather than against a single deterministic environment estimate. For example, \citet{quintero2024stochastic} use variational inference to learn robot signed distance models from noisy distance measurements and correct candidate plans subject to chance-constraints, while \citet{liu2023radius} integrate obstacle probability density over geometric regions to compute collision probability. However, these methods typically represent uncertainty in sensing, obstacle state, or system evolution while relying on deterministic robot swept geometry. Other uncertainty-aware approaches likewise evaluate validity at discrete configurations or trajectory samples~\cite{luders2010chance,meng2022nr}. What remains missing is a swept representation that is simultaneously efficient and continuous over robot motion, while also expressing uncertainty in the learned geometry and allowing it to be combined with uncertainty in perception.

In this work, we propose a probabilistic formulation of swept volume signed distance modeling for uncertainty-aware motion planning.
Our method leverages spherical representations of robot geometry and exploits their geometric properties to efficiently compute both interior and exterior swept distances and to provide gradient supervision that facilitates learning.
We model the swept volume distance function as a probabilistic field that jointly represents geometric occupancy and predictive uncertainty, naturally enabling incorporation of perception and model uncertainty.
Building on this representation, we formulate collision avoidance as a chance constraint within a trajectory  optimization problem to provide probabilistic safety over continuous trajectories.
We demonstrate that the proposed formulation generalizes across multiple robot manipulators, produces tighter and more adaptive safety estimates than post-hoc conformal calibration approaches, naturally incorporates multiple uncertainty sources, and enables efficient trajectory optimization in dynamic environments.

\section{Method}
\label{sec:method}

We study the problem of collision-free motion planning for a robotic manipulator operating under perception uncertainty and approach this problem using stochastic neural swept-volume models. We (i) formulate swept volume representation as a probabilistic signed distance field learning problem, (ii) develop a chance-constrained trajectory optimization problem for motion planning under uncertain obstacle observations, and (iii) extend the framework to a receding-horizon planning scheme that incorporates velocity-level control for dynamic obstacle avoidance in time-varying environments.

\subsection{Preliminaries: Problem Formulation}
\label{sec:problem}

The workspace contains \(n_{\text{obs}}\) obstacles represented as a point cloud.
Each obstacle is defined by its position and radius, such that the obstacle set is given by \(\mathcal{O} = \{(o_i, r_i)\}_{i=1}^{n_\text{obs}}\), where \(o_i\in\mathbb{R}^3\) and \(r_i>0\) denote the center and radius of the \(i\)-th obstacle, respectively.
At perception time, the robot does not have access to the true obstacle centers and instead observes noisy measurements \(\hat{\mathcal{O}} = \{{\hat{o}_i}\}_{i=1}^{n_{\text{obs}}}\).

Let the configuration space of an \(n_d\) manipulator be \(\mathcal{Q} \subset \mathbb{R}^{n_d}\), and let \(\mathcal{V}(q) \subset \mathbb{R}^3\) denote the volume occupied by the robot in the workspace at configuration \(q \in \mathcal{Q}\).
A configuration is collision-free if the robot's occupied volume does not intersect any obstacle in the environment, i.e., \(\mathcal{Q}_\text{free} = \{q\in \mathcal Q\mid \mathcal{V}(q)\cap \mathcal{O} = \emptyset\}\).
Given start and goal configurations \(q_{\text{start}}, q_{\text{goal}} \in \mathcal{Q}_{\text{free}}\), our objective is to compute a continuous trajectory \(\tau:[0,1]\rightarrow \mathcal{Q}_\text{free}\) that remains collision-free with respect to the unknown true environment, despite the perception noise.

\subsection{Preliminaries: Signed Distance Fields and Swept Volume}
\label{sec:prelim}

We begin by introducing the signed distance field (SDF) of a set. Let \(\Omega \subset \mathbb{R}^3\) be a closed set with non-empty interior and boundary \(\partial \Omega\). The signed distance function of a point \(x \in \mathbb{R}^3\) to the set is a mapping \(\mathrm{s}_\Omega: \mathbb{R}^3\rightarrow\mathbb{R}\) defined by
\begin{equation}
  \mathrm{s}_{\Omega}(x) =
  \begin{cases}
    +\mathrm{d}(x, \partial \Omega), & x \notin \Omega, \\
    0, & x \in \partial \Omega, \\
    -\,\mathrm{d}(x, \partial \Omega), & x \in \Omega,
  \end{cases}
\end{equation}
where \(\mathrm{d}(x, \partial \mathcal{A}) = \inf_{y\in\partial\Omega}\|x-y\|_2\) denotes the Euclidean distance from \(x\) to the boundary of \(\Omega\). By construction, \(\mathrm{s}_\Omega\) encodes both the occupancy of \(\Omega\) and the proximity to the boundary, which can be recovered from the zero level set \(\{x\mid \mathrm{s}_\Omega(x)=0\}\). This makes SDF naturally suitable for collision avoidance: a point \(x\) is collision-free with respect to \(\Omega\) if and only if \(\mathrm{s}_\Omega(x) > 0\). Whenever it is differentiable, the signed distance function satisfies the eikonal equation
\begin{equation}
  \|\nabla \mathrm{s}_\Omega(x)\|_2 = 1,
\end{equation}
with its gradients \(\nabla \mathrm{s}_\Omega(x)\) pointing from \(x\) along the direction of steepest increase of the signed distance.

We now apply this construction to the geometry traced out by a moving robot. Given a robot's continuous trajectory in the configuration space \(\tau:[0,1]\rightarrow\mathcal{Q}\), its swept volume is
\begin{equation}
  \mathcal{SV}(\tau) = \bigcup_{t\in [0,1]}\mathcal{V}(\tau(t)) \subset \mathbb{R}^3.
\end{equation}

Specializing the signed distance construction above to \(\Omega = \mathcal{SV}(\tau)\) yields the swept volume signed distance function \(\mathrm{s}_{\mathcal{SV(\tau)}}\). A trajectory \(\tau\) is collision-free with respect to an obstacle point \(x \in \mathbb{R}^3\) if and only if \(\mathrm{s}_{\mathcal{SV(\tau)}}(x) > 0\), reducing continuous-time collision checking along \(\tau\) to a point-wise inequality on a single scalar field.

In practice, the trajectory \(\tau\) is parameterized by a finite vector of decision variables such as waypoints, spline coefficients, or polynomial parameters. In this work we adopt a waypoint formulation with linear interpolation between consecutive configurations, and write
\begin{align}
  &\mathcal{SV}(q_0, q_1) \coloneq \mathcal{SV}(\tau_{q_0, q_1}),\\
  & \tau_{q_0, q_1}(t) = (1-t)q_0 + tq_1,\ t\in [0,1]
\end{align}
so that \(\mathcal{SV}(q_0, q_1)\) denotes the swept volume of the robot traversing in a straight line from \(q_0\) to \(q_1\) in configuration space. This formulation is chosen to be compatible with sampling-based planners, which evaluate collisions along interpolated segments between graph vertices~\cite{kuffner2000rrt, thomason2024motions}, and with trajectory optimizers that represent paths as sequences of such segments. Trajectories with other parameterizations, such as B-splines, can be handled similarly~\cite{joho2024neural}.

\subsection{Ground-Truth Generation via Spherical Representations}
\label{sec:gt}

Training a neural swept volume SDF requires a large dataset of \((\tau, x, \mathrm{s}_{\mathcal{SV}(\tau)}(x))\) triples. However, collecting the data for an articulated manipulator with general mesh is time consuming \cite{joho2024neural, sellan2021swept} and requires numerically sensitive operations, making it difficult to scale to the millions of samples needed to train a network that generalizes across trajectories, query locations, and robot morphologies.

We mitigate this bottleneck by approximating the robot as a union of {$N$} spheres. This leverages their efficiency for collision checking~\cite{thomason2024motions, sundaralingam2023curobo} and replaces the true robot occupancy with a conservative spherized approximation. The signed distance from an exterior query point $x$ to the swept region then admits a closed-form expression
\begin{equation}
  \mathrm{s}_{\hat{\mathcal{SV}}(\tau)}(x) = \min_{\substack{
      t\in[0,1] \\
      i=1,\cdots,N
  }}\mathrm{s}_{\mathcal{B}_{t,i}}(x)
\end{equation}
for \(x \not \in \hat{\mathcal{SV}}(q_0, q_1) \). \(\mathcal{B}_{t,i} = \mathcal{B}(c_i(\tau(t)), r_i)\) is a single sphere with time parameterized center and radius.
This representation also provides direct targets for the spatial gradient of the swept-volume SDF from the Eikonal identity:
\begin{align}
  \nabla \mathrm{s}_{\hat{\mathcal{SV}}(\tau)}(x) & = \frac{x-c^*}{\|x-c^*\|_2} \label{eqn:gradient} \\
  c^* &= c_{i^*}(\tau(t^*)) \\
  t^*, i^* &= \argmin_{\substack{
      t\in[0,1] \\
      i=1,\cdots,N
  }}\mathrm{s}_{\mathcal{B}_{t,i}}(x).
\end{align}

\begin{figure*}
  \centering \includegraphics[width=1.0\linewidth]{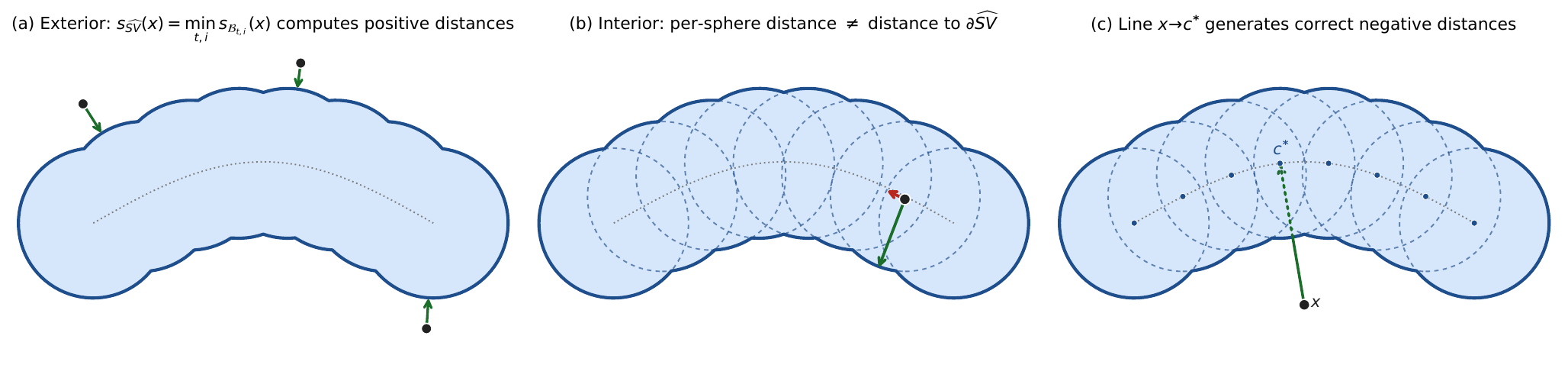}
  \caption{Ground-truth swept-volume SDF computation via spherical representations. The blue region depicts the swept volume of a moving sphere. Solid black dots mark query points; green arrows indicate their signed distances to the swept-volume boundary. In (b), the red arrows show the incorrect distance produced by the naive formulation at an interior point, which measure distance to an individual sphere boundary rather than to the boundary of the union.}
  \label{fig:sdf-illustration}
\end{figure*}

The interior case is more subtle, as when \(x\) lies inside the union, naively taking the maximum of negative per-sphere signed distances gives the distance to the nearest individual sphere boundary rather than to the union's boundary, as shown in Fig.~\ref{fig:sdf-illustration}~(b). Our key observation is that to train a network, interior points need not be directly sampled.  Since the closed-form gradient~\eqref{eqn:gradient} at any exterior \(x\) points along the line from the nearest swept-sphere center \(c^*\) to \(x\), we can trace this ray inward and generate interior samples from ground-truth signed distance known by construction. 
Specifically, given an exterior sample \((x,d)\) with \(d=\mathrm{s}_{\hat{\mathcal{SV}}(\tau)}(x)>0\), we sample an interior offset \(\beta\in(0,r_{i^*})\) and step along the negative distance gradient by \(d+\beta\):
\begin{equation}
    x' = x-(d+\beta)\nabla_x \mathrm{s}_{\hat{\mathcal{SV}}(\tau)}(x).
\end{equation}
The resulting point lies inside the corresponding swept sphere at depth \(\beta\), with signed distance \(-\beta\). Varying \(\beta\) generates dense interior samples from exterior points without directly evaluating an interior union distance.
Figure~\ref{fig:sdf-illustration}(c) illustrates such construction. Using this formulation, our implementation using jax \cite{jax2018github} generates millions of training points in seconds.

\subsection{Learning swept volume SDF from a Probabilistic View}

The constructions in Sections~\ref{sec:prelim} and~\ref{sec:gt} treat the swept volume signed distance function as a deterministic geometric object given a trajectory. In deployment, however, obtaining the exact form is intractable. The SDF \(\mathrm{s}_{\hat{\mathcal{SV}}(\tau)}\) is approximated with a neural network at planning time; obstacle locations against which the swept volume is checked come from noisy perception. A common approach is to inflate the collision constraint using a fixed global safety margin. However, such a margin can be overly conservative in regions where the model is accurate, while still being insufficient in regions with large model approximation error or high perception uncertainty. To make these errors explicit, we adopt a probabilistic view and treat the swept volume signed distance at a query point as Normally-distributed random variable. We note that this assumption is a working model rather than a claim about the true distribution of modeling error.

Formally, given a sample \((\tau_k, x_k, s_k, g_k)\) of a trajectory, a query point, the corresponding ground-truth signed distance \(s_k = \mathrm{s}_{\hat{\mathcal{SV}}(\tau_k)}(x_k)\) and ground-truth spatial gradient \(g_k = \nabla_x \mathrm{s}_{\hat{\mathcal{SV}}(\tau_k)}(x_k)\), the loss terms are computed by
\begin{align}
  & \mathcal{L}_{\mathrm{nll}}(\theta) = \frac{\big(s_k - \mu_\theta(x_k, \tau_k)\big)^2}{2\,\sigma^2_\theta(x_k, \tau_k)} + \tfrac{1}{2}\log \sigma^2_\theta(x_k, \tau_k), \\
  & \mathcal{L}_{\mathrm{grad}}(\theta) = \big\| \nabla_x \mu_\theta(x_k, \tau_k) - g_k \big\|_2^2,
\end{align}
where \(\mu_\theta\) and \(\sigma^2_\theta\) are the predicted mean and variance of the network with parameters \(\theta\). The negative log-likelihood term \(\mathcal{L}_{\mathrm{nll}}\) encourages the network to predict large \(\sigma^2\) where its mean predictions are unreliable, and small \(\sigma^2\) where they are accurate. The gradient term \(\mathcal{L}_{\mathrm{grad}}\) supervises the spatial gradient of the predicted mean against the ground truth to satisfy the property of the SDF. The full training objective combines both terms with weighting term $\lambda_{\text{grad}}$:
\begin{equation}
  \mathcal{L}(\theta) =  \mathcal{L}_{\mathrm{nll}}(\theta) + \lambda_{\mathrm{grad}}\, \mathcal{L}_{\mathrm{grad}}(\theta)
\end{equation}

\subsection{Chance-Constrained Trajectory Optimization} \label{sec:cc}

We now describe how the learned swept-volume representation is integrated into manipulator motion planning through chance-constrained trajectory optimization. Our goal is to compute a collision-free trajectory while accounting for multiple sources of uncertainty, including perception uncertainty and modeling uncertainty in the learned swept volume field. Given start and goal configurations $q_{\text{start}}$ and $q_{\text{goal}}$, we seek a sequence of intermediate waypoints $q_{0:T}$ such that each trajectory segment satisfies a probabilistic collision-safety requirement. Formally, we consider the following chance-constrained optimization problem:
\begin{align} \label{opt:chance-constrained}
\setlength{\abovedisplayskip}{1pt}
\setlength{\belowdisplayskip}{1pt}
    \min_{q_{0:T}} \hspace{0.3cm} & \mathcal{C}_{\text{task}} \left( q_{0:T}\right)\nonumber\\ 
    \textbf{s.t.} \hspace{0.3cm}  & q_0 = q_{\text{start}}, q_T = q_{\text{goal}}, \nonumber \\
    & \underline{q}_{\text{lim}} \le q_t \le \bar{q}_{\text{lim}},\\
    & \operatorname{Pr} \left( \mathcal{SV}(q_t, q_{t+1}) \cap \mathcal{O} = \emptyset\right) \geq 1-\delta \nonumber \\
    & \hspace{4cm} \forall t=0,\cdots,T-1.
\nonumber
\end{align}
where $\mathcal{C}_{\text{task}}$ denotes a user-defined trajectory objective, such as path length or smoothness, and $\underline{q}_{\text{lim}}, \bar{q}_{\text{lim}}$ represent the joint limits. The parameter $\delta$ specifies the maximum allowable probability of collision for each trajectory segment.

In practice, to mitigate local minima, we solve~\eqref{opt:chance-constrained} using the penalty-based particle trajectory optimization framework SPaSM~\cite{chen2025differentiable}. Specifically, we sample a batch of initial trajectories and optimize best candidates in parallel. The resulting objective for each candidate takes the form
\begin{align}\label{eq:cost}
\min_{q_{1:T-1}} \quad
& w_{\mathrm{task}}
\mathcal{C}_{\mathrm{task}}(q_{0:T})
\nonumber\\
+\ & w_{\mathrm{collision}}
\sum_{i=1}^{n_{\mathrm{obs}}}
\sum_{t=0}^{T-1}
\mathcal{C}_{\mathrm{safe}}
(q_t,q_{t+1},\hat{o}_i,P_i, \theta),
\end{align}
where $\hat{o}_i$ is the perceived position of obstacle $i$ with perception uncertainty $P_i$, represented as a covariance matrix. To quantify probabilistic safety, we define the collision penalty
\begin{align}
& \mathcal{C}_{\text{safe}}(q_t, q_{t+1}, \hat{o}_i, P_i, \theta ) = \max\Big(0, \nonumber\\
& -\mu_\theta(\hat{o}_i, \tau_{\cdot, \cdot}) + \alpha \sqrt{\sigma^2_\theta(\hat{o}_i, \tau_{q_t, q_{t+1}}) + {n_{\theta, i}^{\top}P_i n_{\theta,i}}}\ \Big), \\
& n_{\theta,i} = \nabla_{\hat{o}_i}\mu_\theta(\hat{o}_i, \tau_{q_t, q_{t+1}}),\ \alpha = \Phi^{-1}(1-\delta).
\end{align}
Here $\Phi^{-1}$ is the inverse cumulative distribution function of the standard normal distribution, which translates user-specified chance tolerance $\delta$ to the corresponding confidence multiplier $\alpha$. The term $n_{\theta,i}^{\top} P_i n_{\theta,i}$ corresponds to the variance in the signed-distance estimate induced by the predicted obstacle-position uncertainty. Intuitively, the obstacle covariance is projected onto the local distance gradient direction, yielding the effective uncertainty along the collision boundary normal via a  first order Taylor expansion. Assuming independence between perception and model uncertainty, this propagated variance is added to the predictive variance $\sigma_\theta^2$; the square root converts their combined variance into a standard deviation, which is then scaled by confidence multiplier $\alpha$ to form the probabilistic safety margin. In the isotropic case where $P_i = \sigma_p^2 \mathbb{I}$, the term simplifies to $\sigma^2_p$ with $\|n_{\theta,i}\|_2 = 1$ for a signed distance field.

This formulation provides a unified treatment of both geometric modeling uncertainty and perception uncertainty within a differentiable optimization objective. Rather than relying on a conservative global safety margin, the planner adaptively scales the safety buffer according to the local uncertainty predicted by the learned swept-volume model.

\subsection{Receding-Horizon Trajectory Planning}
The chance-constrained formulation in Section~\ref{sec:cc} produces a single open-loop trajectory in static environments. In dynamic environments, robot must react to obstacle motions and evolving perception as well as prediction uncertainties. To handle this setting, we adopt a receding-horizon formulation in which the planner repeatedly solves a short-horizon chance-constrained optimization problem, executes the first control action, incorporates updated observations, and replans.

The trajectory optimization is formulated as a sequence of linked optimization problems as (\ref{opt:chance-constrained});
\begin{align} \label{eqn:mpc}
    \min_{u_{0:H}} \quad
    & \mathcal{C}_{\mathrm{task}}(q_{0:H+1}, q_T) \quad \textbf{s.t.} \; \; q_{t+1} = f(q_t, u_t)
\end{align}
where $u_t \in \mathbb{R}^{n_d}$ denotes the joint-velocity command at time step $t$, $f(q_t, u_t)$ is the time-discretized configuration-space dynamics model, and $H$ is the planning horizon. The cost function defines the desired behavior to complete the task, such as progression toward the goal configuration.

\section{Experimental Results}
\label{sec:experiments}

This section presents four sets of analyses to evaluate the proposed framework. We first compare probabilistic and deterministic swept volume SDF models in terms of prediction accuracy and uncertainty quantification. We then evaluate their ability to correct potentially colliding RRT trajectories into safe continuous trajectories using MotionBenchMaker (MBM)~\cite{chamzas2021motionbenchmaker}, under varying levels of perception noise. Next, we  assess the effectiveness of the proposed motion planning formulation in a cluttered environment containing 150 spherical obstacles with noisy observations. Finally, we evaluate the system in dynamic environments using a receding-horizon control scheme with online re-planning. 

Experiments are performed on a machine with AMD Ryzen Threadripper PRO 5965WX 24-Cores CPUs and an NVIDIA RTX 4090 GPU.

\subsection{Swept Volume Learning and Uncertainty Quantification}

We first evaluate the learning performance of the proposed probabilistic swept volume SDF. We study (i) whether the learned mean field $\mu_\theta$ accurately approximates swept-volume SDF across different robot morphologies, and (ii) whether the learned variance field $\sigma_\theta^2$ provides a calibrated, query-dependent estimate of predictive uncertainty.

 We conduct experiments on three manipulators with distinct kinematic structures: a 7-DOF Franka Panda, a 6-DOF Universal Robots UR5, and an 8-DOF Fetch arm. For each robot, we generate a dataset of 1 million linear joint-space trajectories sampled uniformly within the joint limits along with 512 query points sampled uniformly from the robot’s reachable workspace. Ground-truth signed distances are computed using the method described in Section~\ref{sec:gt}. For every experiments, we train the same 12-layer, 1024 wide MLP with default \texttt{torch} AdamW optimizer.

 We compare the proposed probabilistic model trained using a negative log-likelihood (NLL) loss, against a deterministic baseline trained using mean squared error (MSE) loss on the same architecture and dataset. Both models include gradient loss, such that the regression term is their only difference. To evaluate uncertainty quantification, we further perform a coverage analysis. For the MSE baseline, we apply a post-hoc conformal-style calibration procedure in which test-set prediction errors are modeled as samples from a Gaussian distribution, and the empirical standard deviation is used as a fixed, query-independent uncertainty estimate.
    
Table~\ref{tab:learning-comparison} summarizes the evaluation results. Across all robot platforms, models trained with the NLL loss consistently achieve lower prediction error than those trained with the MSE loss, with the largest improvement observed for the Panda robot, whose geometry is more complex. In addition, the NLL-trained models generally predict smaller standard deviations while achieving higher empirical coverage across the 1-, 2-, and 3-standard-deviation intervals. indicating that the learned variance provides a more reliable estimate of modeling uncertainty. In contrast, achieving a comparable level of safety with an MSE-trained model would require a larger global safety buffer, leading to a more conservative over-approximation. Figure~\ref{fig:reconstruction} visualizes examples of the reconstructed swept volumes for all three robots.

\begin{table}[b]
    \centering
    \begin{tabular}{ccccccc}
    \toprule
     \multirow{2}{*}{Robot} & \multirow{2}{*}{Loss}  & \multirow{2}{*}{\makecell{Mean\\ Error $\downarrow$}} & \multirow{2}{*}{\makecell{Mean\\ Std}} & \multicolumn{3}{c}{Coverage}  \\
     & & & & $\alpha=1$ & $\alpha=2$ & $\alpha=3$ \\\midrule
     \multirow{2}{*}{Panda} & NLL & \textbf{3.4} & 4.4 & 78.4 & 96.5 & 99.4 \\ %
         & MSE & 4.7 & 5.8 & 73.4 & 91.2 & 96.5 \\ \hline 
     \multirow{2}{*}{UR5} & NLL & \textbf{3.7} & 4.9 & 81.4 & 97.6 & 99.5 \\ %
         & MSE & 4.4 & 5.3  & 72.2 & 90.8  & 96.4 \\ \hline 
     \multirow{2}{*}{Fetch} & NLL & \textbf{4.2} & 5.5 & 85.9 & 97.9 & 99.5 \\
     & MSE & 4.9 & 7.6 & 79.8 & 91.3 & 96.2 \\ \bottomrule
    \end{tabular}
    \caption{Learning and uncertainty quantification performance of the probabilistic (NLL) and deterministic (MSE) swept-volume SDF models across three robots. Mean error and std are in unit [mm]. }
    \label{tab:learning-comparison}
\end{table}

\subsection{Continuous Trajectory Refinement}
\label{sec:continuous-trajectory-refinement}
We now evaluate the ability of our neural swept volume model to refine discrete and potentially in-collision trajectories. Sampling-based planners typically represent a solution trajectory as a sequence of discrete waypoints connected by interpolated motions and collision-checked only at a finite set of configurations. Consequently, a trajectory considered collision-free under discrete collision checking may still collide with obstacles along its continuous execution.

We therefore apply our neural swept volume model to refine trajectories generated by sampling-based planners and evaluate its ability to recover collision-free continuous motions. Specifically, we generate waypoint-based trajectories using RRT-Connect~\cite{kuffner2000rrt, thomason2024motions} on MotionBenchMaker (MBM)~\cite{chamzas2021motionbenchmaker} and use the resulting trajectories to initialize the optimization problem~\eqref{opt:chance-constrained}, with the intermediate waypoints treated as optimization variables. We evaluate both the deterministic model trained with MSE loss and the probabilistic model trained with NLL loss, using only the predicted mean ($\alpha=0$) for the latter, focusing on the ability of the learned swept volume representations to recover collision-free continuous trajectories from discretely validated planner outputs.

\begin{table*}[hbtp]
\centering\small
\setlength{\tabcolsep}{3pt}
\begin{tabular}{llcccccccccccc}
\toprule
& & \multicolumn{4}{c}{$\sigma_p = 0$\,m} & \multicolumn{4}{c}{$\sigma_p = 0.01$\,m} & \multicolumn{4}{c}{$\sigma_p = 0.02$\,m} \\
\cmidrule(lr){3-6}\cmidrule(lr){7-10}\cmidrule(lr){11-14}
Task & Method & SR $\uparrow$ & CR $\downarrow$ & Time $\downarrow$ & Len $\downarrow$ & SR $\uparrow$ & CR $\downarrow$ & Time $\downarrow$ & Len $\downarrow$ & SR $\uparrow$ & CR $\downarrow$ & Time $\downarrow$ & Len $\downarrow$ \\
\midrule
\multirow{3}{*}{Bookshelf (small)}
 & RRT-Connect & 25 & 75 & -- & 4.2$\pm$0.7 & 27 & 71 & -- & 4.6$\pm$0.9 & 33 & 56 & -- & 4.6$\pm$0.8 \\
 & SV (MSE) & \snd{32} & \snd{68} & 1.0$\pm$0.7 & 4.7$\pm$1.3 & \snd{37} & \snd{61} & 1.0$\pm$0.6 & 4.9$\pm$1.1 & \snd{38} & \snd{51} & 1.0$\pm$0.6 & 5.0$\pm$1.4 \\
 & SV (NLL) & \best{35} & \best{65} & 1.0$\pm$0.7 & 5.3$\pm$1.5 & \best{39} & \best{59} & 1.0$\pm$0.6 & 5.6$\pm$1.4 & \best{46} & \best{43} & 1.0$\pm$0.6 & 6.4$\pm$1.8 \\
\midrule
\multirow{3}{*}{Bookshelf (tall)}
 & RRT-Connect & 33 & 67 & -- & 4.5$\pm$0.6 & 37 & 63 & -- & 4.6$\pm$0.6 & 40 & 58 & -- & 4.7$\pm$0.7 \\
 & SV (MSE) & \snd{43} & \snd{57} & 1.5$\pm$0.9 & 4.6$\pm$0.7 & \snd{43} & \snd{57} & 1.5$\pm$0.9 & 4.8$\pm$0.9 & \snd{41} & \snd{57} & 1.6$\pm$0.9 & 5.1$\pm$1.1 \\
 & SV (NLL) & \best{48} & \best{52} & 1.5$\pm$0.9 & 5.3$\pm$1.2 & \best{51} & \best{49} & 1.5$\pm$0.9 & 5.7$\pm$1.4 & \best{56} & \best{42} & 1.6$\pm$0.9 & 6.3$\pm$1.7 \\
\midrule
\multirow{3}{*}{Bookshelf (thin)}
 & RRT-Connect & 20 & 80 & -- & 4.4$\pm$0.6 & 22 & 78 & -- & 4.6$\pm$0.6 & 38 & 59 & -- & 4.4$\pm$0.6 \\
 & SV (MSE) & \snd{67} & \snd{33} & 3.1$\pm$1.4 & 5.1$\pm$1.1 & \snd{74} & \snd{26} & 3.3$\pm$1.4 & 5.4$\pm$1.1 & \snd{74} & \snd{23} & 3.5$\pm$1.5 & 6.2$\pm$1.7 \\
 & SV (NLL) & \best{81} & \best{19} & 3.1$\pm$1.4 & 6.6$\pm$1.6 & \best{81} & \best{19} & 3.3$\pm$1.5 & 7.0$\pm$1.7 & \best{87} & \best{10} & 3.5$\pm$1.5 & 7.4$\pm$1.6 \\
\midrule
\multirow{3}{*}{Cage}
 & RRT-Connect & 7 & 93 & -- & 6.0$\pm$0.6 & 13 & 85 & -- & 6.3$\pm$1.2 & 14 & 57 & -- & 6.6$\pm$1.8 \\
 & SV (MSE) & \snd{45} & \snd{55} & 2.3$\pm$1.1 & 7.9$\pm$2.7 & \snd{39} & \snd{59} & 2.7$\pm$1.2 & 9.6$\pm$2.7 & \snd{30} & \snd{41} & 3.5$\pm$1.6 & 11.8$\pm$3.8 \\
 & SV (NLL) & \best{58} & \best{42} & 2.4$\pm$1.1 & 9.8$\pm$2.9 & \best{54} & \best{44} & 2.7$\pm$1.2 & 11.6$\pm$2.8 & \best{52} & \best{19} & 3.5$\pm$1.6 & 15.4$\pm$4.9 \\
\midrule
\multirow{3}{*}{Table Pick}
 & RRT-Connect & 47 & 53 & -- & 4.6$\pm$0.6 & 61 & 37 & -- & 4.6$\pm$0.6 & 62 & 25 & -- & 4.7$\pm$0.6 \\
 & SV (MSE) & \snd{87} & \snd{13} & 1.0$\pm$0.4 & 4.9$\pm$0.9 & \snd{76} & \snd{22} & 1.1$\pm$0.5 & 5.0$\pm$0.9 & \snd{76} & \snd{11} & 1.3$\pm$0.6 & 5.8$\pm$1.7 \\
 & SV (NLL) & \best{95} & \best{5} & 1.0$\pm$0.4 & 5.3$\pm$1.1 & \best{92} & \best{6} & 1.1$\pm$0.5 & 5.7$\pm$1.3 & \best{81} & \best{6} & 1.3$\pm$0.6 & 6.7$\pm$1.9 \\
\midrule
\multirow{3}{*}{Table Under Pick}
 & RRT-Connect & 23 & 77 & -- & 7.1$\pm$1.8 & 36 & 63 & -- & 6.6$\pm$1.8 & 47 & 41 & -- & 6.6$\pm$1.5 \\
 & SV (MSE) & \snd{80} & \snd{20} & 2.3$\pm$0.8 & 6.9$\pm$2.1 & \snd{87} & \snd{12} & 2.3$\pm$0.9 & 7.2$\pm$2.0 & \snd{76} & \snd{12} & 2.6$\pm$0.9 & 8.5$\pm$2.4 \\
 & SV (NLL) & \best{92} & \best{8} & 2.3$\pm$0.8 & 8.2$\pm$2.2 & \best{94} & \best{5} & 2.4$\pm$0.9 & 8.7$\pm$2.1 & \best{85} & \best{3} & 2.6$\pm$0.9 & 10.8$\pm$2.5 \\
\bottomrule
\end{tabular}
\caption{RRT-Connect trajectory refinement results on MBM under different perception-noise levels $\sigma_p$. SR: success rate [\%], CR: collision rate [\%], Time: planning time [s], Len: path length. \best{Best} and \snd{second best} per column within each task. Planning time and path length are unranked: both swept-volume variants share the same optimization budget, and shorter paths here reflect less conservative avoidance rather than better performance.}
\label{tab:mbm_rrt_spasm_results}
\end{table*}

We conduct 100 trials for each task and perception-noise level. Uncertainty in environment perception is modeled by perturbing each observed obstacle position with isotropic Gaussian noise such that $\hat{o}_i = o_i + \epsilon_i,\ \epsilon_i \sim \mathcal{N}(0,\sigma_p^2\mathbb{I}),$
with $\sigma_p\in \{0.0,0.01,0.02\}$, while evaluating collisions and task success against the complete ground-truth obstacle geometry.
For both swept volume models, we optimize the intermediate trajectory waypoints for 50 iterations using the cost function defined in Equation~\ref{eq:cost} given the ground truth noise covariance, with $w_{\text{task}}=0$ and $w_{\text{safe}}=1$. We report success rate, collision rate, planning time, and path length, with path length averaged only over successful trials.

\begin{figure}[t]
    \centering
    \vspace{-0.2cm}
    \begin{subfigure}[b]{0.23\textwidth}
        \centering
        \includegraphics[trim={0cm 0cm 0cm 0cm},clip, width=1\textwidth]{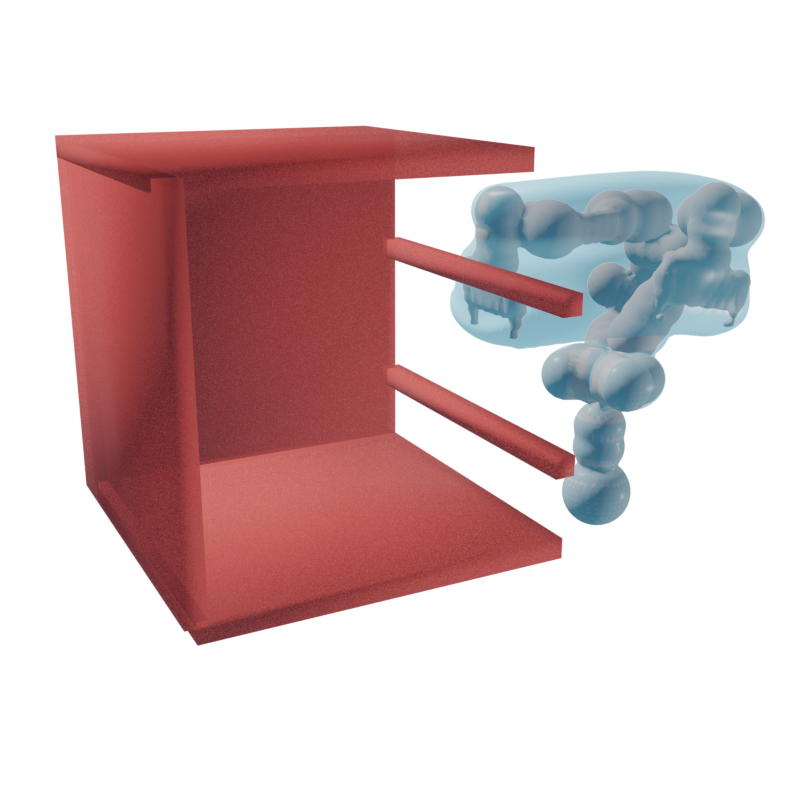}
    \end{subfigure}
    \begin{subfigure}[b]{0.23\textwidth}
        \centering
        \includegraphics[trim={0cm 0cm 0cm 0cm},clip, width=1\textwidth]{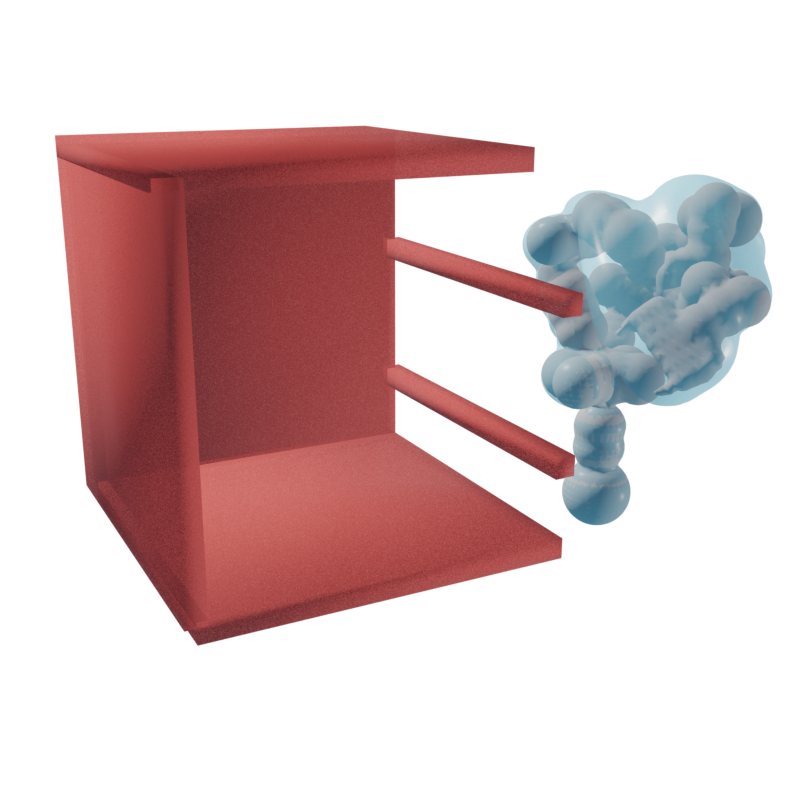}
    \end{subfigure}
    \par\vspace{-1.0cm}
    \begin{subfigure}[b]{0.23\textwidth}
        \centering
        \includegraphics[trim={0cm 0cm 0cm 0cm},clip, width=1\textwidth]{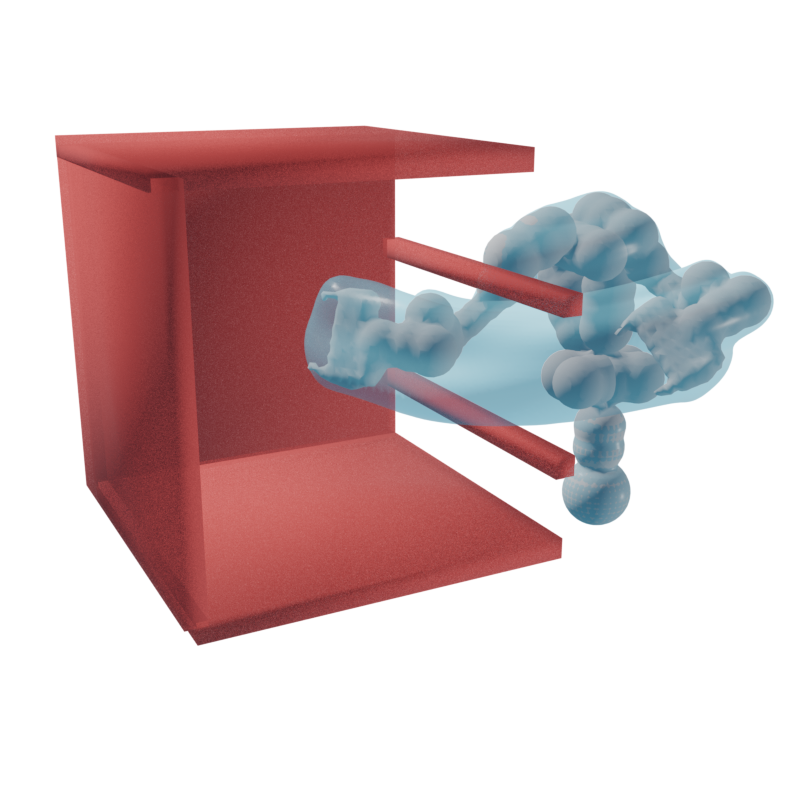}
    \end{subfigure}
    \begin{subfigure}[b]{0.23\textwidth}
        \centering
        \includegraphics[trim={0cm 0cm 0cm 0cm},clip, width=1\textwidth]{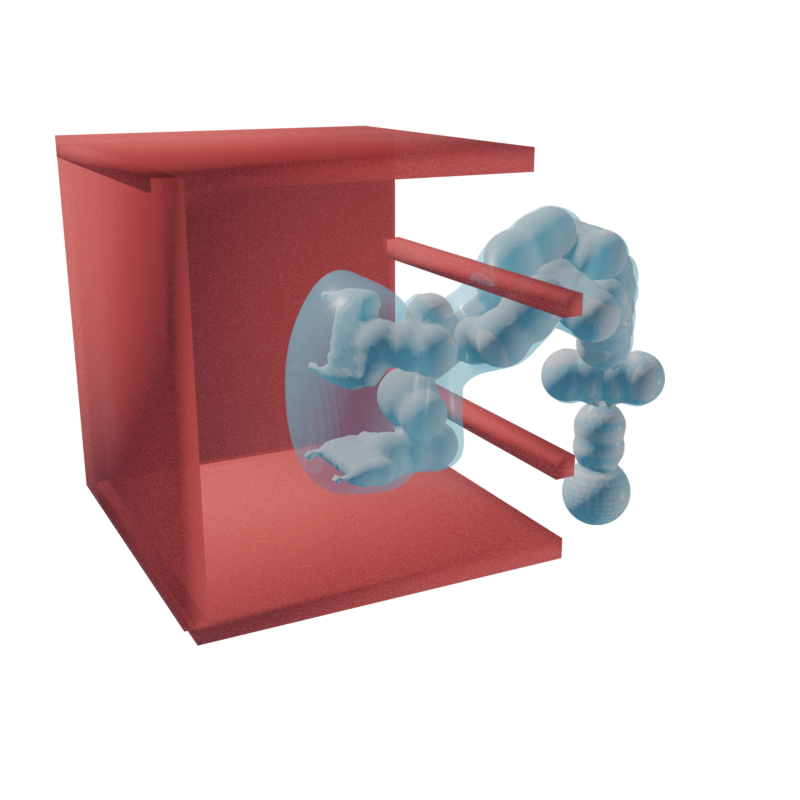}
    \end{subfigure}
    \vspace{-0.6cm}
    \caption{Example planned trajectory in the Cage (red) environment and the corresponding learned swept volume reconstruction (transparent blue). The robot motion is shown sequentially from left to right and top to bottom.}
    \label{fig:mbm}
\end{figure}

Table~\ref{tab:mbm_rrt_spasm_results} summarizes the results, with the original RRT-Connect trajectories included for reference. Although RRT-Connect finds trajectories that are collision-free under discrete checks, many fail continuous collision validation against the full environment geometry. Refinement using either learned swept volume model substantially improves success across most tasks and noise levels. In particular, the NLL model achieves a higher success rate and lower collision rate than the MSE model. An example of the planned trajectory in the Cage environment is shown in Figure~\ref{fig:mbm}.

\subsection{Uncertainty-Aware Path Planning}
While MBM provides diverse and challenging planning problems, many are designed to contain narrow passages, making it difficult to isolate the effect of uncertainty awareness from the effect of geometric clearance. We thus consider a Franka Panda arm tasked with moving from a start pose to a goal pose while avoiding collisions to 150 randomly placed spherical obstacles, and use the same perception-noise model and ground-truth evaluation criterion as in Section~\ref{sec:continuous-trajectory-refinement}.

We compare a straight-line trajectory, CuRobo~\cite{sundaralingam2023curobo}, RRT-Connect~\cite{kuffner2000rrt}, and an uncertainty-aware optimization method CCIKOPT~\cite{quintero2024stochastic}. CCIKOPT learns a distribution over signed distances between the robot and sensed environment points from repeated noisy distance samples during training, and incorporates the prediction into a chance-constrained inverse-kinematics optimization. We additionally evaluate both our  deterministic MSE method, and the uncertainty-aware neural NLL model at different uncertainty margins $\alpha\in\{0,1,2\}$. For both swept volume models, we optimize the intermediate trajectory waypoints for 50 iterations using the cost function defined in Equation~\ref{eq:cost} with $\mathcal{C}_{\mathrm{task}}(q_{0:T})=\sum_{t=1}^T\|q_t - q_{t-1}\|_2$, $w_{\text{task}}=0.01$ and $w_{\text{safe}}=1$.

\begin{table*}[hbtp]
\centering\small
\setlength{\tabcolsep}{4pt}
\begin{tabular}{lcccccccccccc}
\toprule
& \multicolumn{4}{c}{$\sigma_p = 0$\,m} & \multicolumn{4}{c}{$\sigma_p = 0.01$\,m} & \multicolumn{4}{c}{$\sigma_p = 0.02$\,m} \\
\cmidrule(lr){2-5}\cmidrule(lr){6-9}\cmidrule(lr){10-13}
Method & SR $\uparrow$ & CR $\downarrow$ & Time $\downarrow$ & Len $\downarrow$ & SR $\uparrow$ & CR $\downarrow$ & Time $\downarrow$ & Len $\downarrow$ & SR $\uparrow$ & CR $\downarrow$ & Time $\downarrow$ & Len $\downarrow$ \\
\midrule
Straight-line & 17 & 83 & -- & 4.8$\pm$1.1 & 17 & 83 & -- & 4.8$\pm$1.1 & 17 & 83 & -- & \best{4.8$\pm$1.1} \\
CuRobo & 32 & 32 & \snd{0.9$\pm$0.7} & 5.2$\pm$1.5 & 32 & 27 & \best{0.9$\pm$0.8} & 5.8$\pm$1.9 & 24 & 31 & \best{1.0$\pm$0.9} & 5.1$\pm$1.3 \\
RRT-Connect & 60 & 40 & \best{0.5$\pm$0.6} & 7.8$\pm$3.8 & 54 & 46 & 3.0$\pm$1.9 & 5.1$\pm$2.0 & 32 & 68 & \snd{1.8$\pm$1.8} & 5.4$\pm$2.0 \\
\midrule
CCIKOPT ($\alpha$=0) & 16 & 7 & 11.7$\pm$3.3 & \best{4.0$\pm$0.8} & 11 & \best{0} & 13.8$\pm$4.5 & \snd{4.5$\pm$1.0} & 6 & \best{0} & 13.9$\pm$4.2 & \snd{4.9$\pm$0.8} \\
CCIKOPT ($\alpha$=1) & 16 & 7 & 11.5$\pm$3.3 & \best{4.0$\pm$0.8} & 10 & \snd{1} & 17.3$\pm$5.6 & \snd{4.5$\pm$1.0} & 7 & \best{0} & 13.8$\pm$4.1 & 5.7$\pm$2.2 \\
CCIKOPT ($\alpha$=2) & 13 & 7 & 12.0$\pm$3.4 & \snd{4.2$\pm$0.7} & 9 & \best{0} & 14.6$\pm$4.9 & \best{4.3$\pm$1.0} & 3 & \best{0} & 13.7$\pm$4.2 & 5.5$\pm$0.8 \\
\midrule
SV (MSE) & \best{91} & 7 & 1.5$\pm$0.3 & 5.9$\pm$1.2 & 84 & 12 & \snd{1.5$\pm$0.1} & 5.9$\pm$1.3 & 70 & 27 & \snd{1.5$\pm$0.1} & 5.8$\pm$1.3 \\
SV (NLL, $\alpha$=0) & \snd{89} & 8 & 1.5$\pm$0.4 & 6.2$\pm$1.4 & 83 & 14 & \snd{1.5$\pm$0.1} & 6.1$\pm$1.3 & \snd{78} & 20 & \snd{1.5$\pm$0.1} & 6.1$\pm$1.4 \\
SV (NLL, $\alpha$=1) & \best{91} & \snd{3} & 1.5$\pm$0.1 & 6.8$\pm$1.6 & \best{89} & 5 & \snd{1.5$\pm$0.1} & 6.8$\pm$1.6 & \best{81} & \snd{9} & \snd{1.5$\pm$0.1} & 6.7$\pm$1.6 \\
SV (NLL, $\alpha$=2) & 83 & \best{0} & 1.5$\pm$0.1 & 7.0$\pm$1.5 & \snd{86} & \best{0} & \snd{1.5$\pm$0.1} & 7.1$\pm$1.7 & 69 & \best{0} & \snd{1.5$\pm$0.1} & 7.2$\pm$1.9 \\
\bottomrule
\end{tabular}
\caption{Static obstacle planning results under different perception-noise levels $\sigma_p$. Swept Volume (SV) indicates our approach. SR: success rate [\%], CR: collision rate [\%], Time: planning time [s], Len: path length. \best{Best} and \snd{second best} per column.}
\label{tab:static_obstacle_results}
\end{table*}

\begin{figure}[t] %
    \centering
    \begin{subfigure}[b]{0.22\textwidth}
        \centering
        \includegraphics[trim={0cm 0cm 0cm 0.5cm},clip, width=1\textwidth]{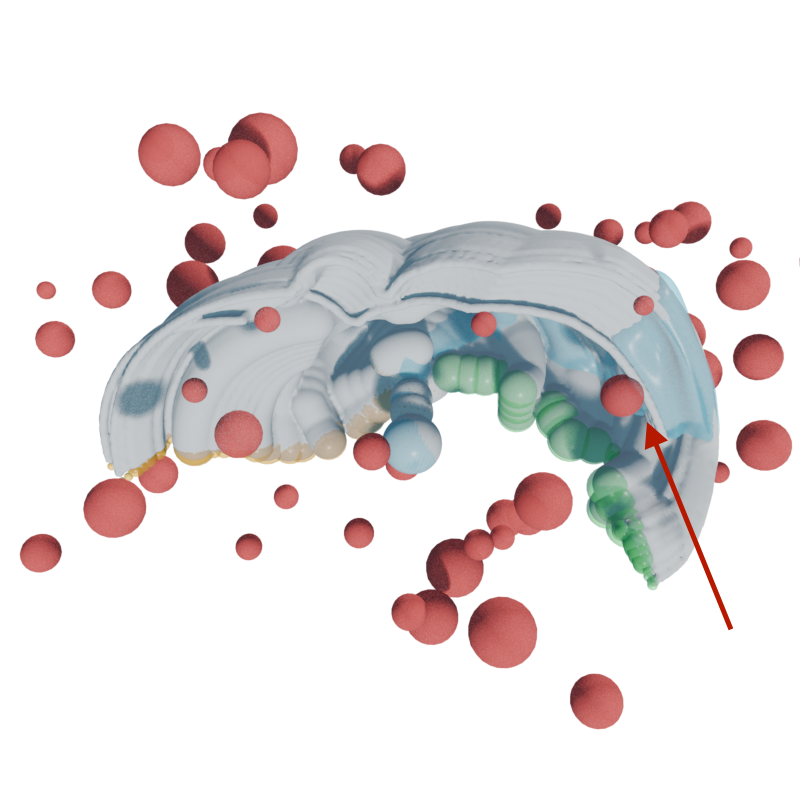}
    \end{subfigure}
    \begin{subfigure}[b]{0.22\textwidth}
        \centering
        \includegraphics[trim={0cm 0cm 0cm 0.5cm},clip, width=1\textwidth]{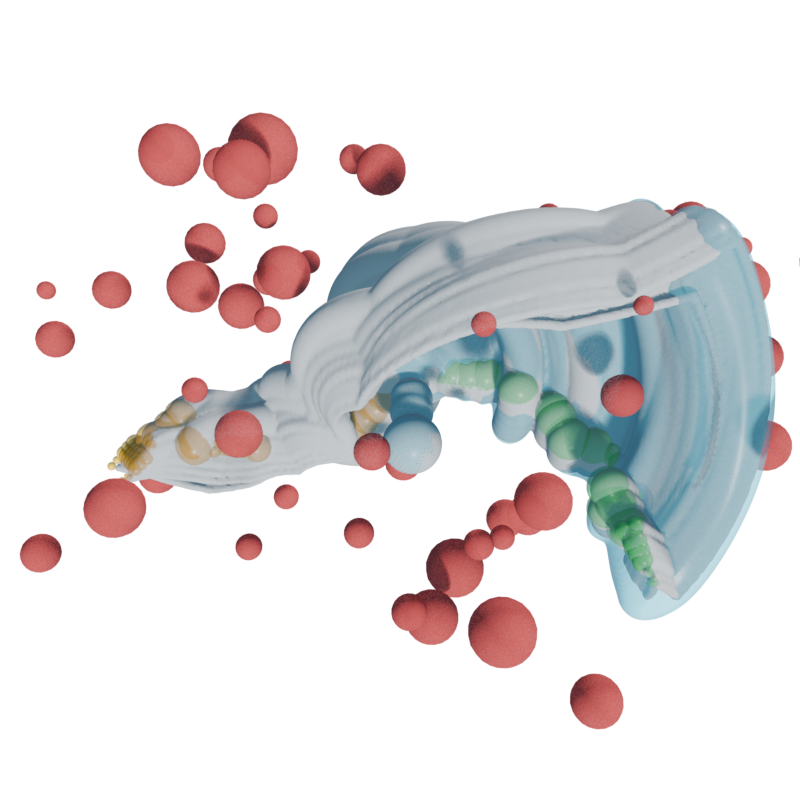}
    \end{subfigure}
    \par
    \vspace{-1em}
    \begin{subfigure}[b]{0.44\textwidth}
        \centering
        \hspace*{-0.05\linewidth}%
        \includegraphics[width=1.10\linewidth]{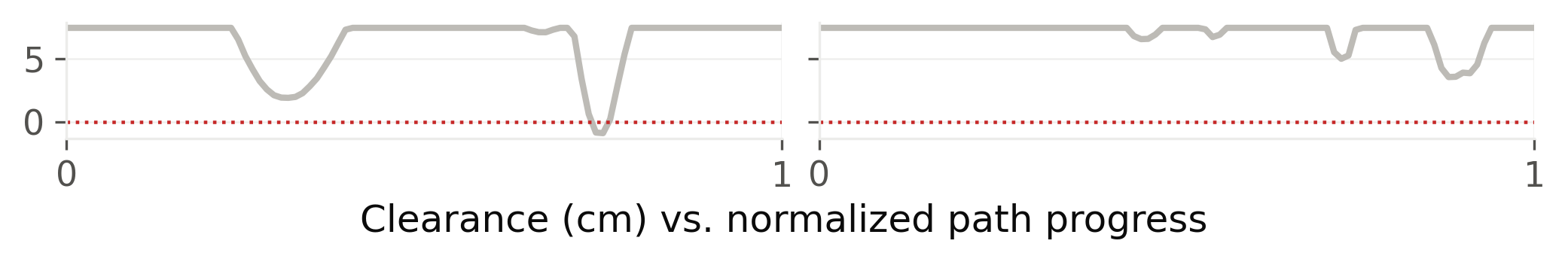}
    \end{subfigure}

    \caption{An illustrative comparison of the paths planned from start (orange) to goal (green). Only the obstacles (red) near the planned trajectory are visualized for clarity of presentation. \textbf{Left:} MSE-trained model. The red arrow marks a collision. The learned swept volume for the colliding segment (transparent blue) under-approximates the true occupied region, causing a collision. \textbf{Right}: NLL-trained model with a \(1\sigma\) uncertainty buffer. Accounting for predictive uncertainty yields a more conservative swept-volume estimate and a collision-free trajectory in critical regions.}
    \label{fig:static}
\end{figure}

Table~\ref{tab:static_obstacle_results} summarizes the results. Explicitly accounting for uncertainty substantially improves robustness to noisy perception. As the observation noise increases, RRT-Connect's collision rate increases considerably. CuRobo maintains relatively low collision rates and short planning times, although its success rates are lower in this setting, likely because it also optimizes additional trajectory-quality objectives (e.g.,  smoothness) beyond collision avoidance. CCIKOPT, as an uncertainty-aware method, also maintains low collision rates across perception-noise levels, but at the cost of lower success rates and longer planning times. At $\sigma_p=0$, its collision rate remains unchanged across risk levels because its stochastic distance model relies on repeated noisy observations to capture uncertainty, which disappear in the absence of perception noise. The deterministic MSE model performs well under low noise, but its collision rate increases as perception noise increases. In contrast, incorporating the uncertainty prediction of the NLL model allows the planner to trade path efficiency for safety. In particular, the NLL model with $\alpha=1$ achieves the highest success rate while maintaining lowest collision rates. Increasing the uncertainty level further eliminates collisions entirely across all three noise levels, at the cost of lower success rates and  longer paths. These results demonstrate that the probabilistic swept volume formulation provides a mechanism for trading path efficiency and feasibility for increased robustness to perception and model uncertainty. Figure~\ref{fig:static} demonstrates this advantage.

\subsection{Planning around Dynamic Obstacles}

We now consider a scenario where a robot manipulator operates in an environment with dynamic obstacles. To introduce uncertainty and unpredictable obstacle dynamics into the environment, we design obstacles to follow sinusoidal trajectories nearby the robot. At planning time, the robot observes only noisy estimates of the current obstacle positions and uses their motion history to infer future states. Specifically, it employs a Kalman Filter (KF) \cite{ribeiro2004kalman} to estimate future obstacle positions and their associated uncertainty, represented as covariance matrices. We then apply the receding-horizon formulation defined in~\eqref{eqn:mpc} and solve the resulting optimization problem using IPOPT \cite{wachter2006implementation}. We use the cost function $\mathcal{C}_\text{task} (q_{0:H+1}, q_T) = \|q_{H+1} - q_T\|_2^2$ and formulate obstacle avoidance as a hard constraint.

We evaluate the proposed method over 100 simulated scenarios with varying obstacle initial conditions and motion parameters, including amplitude, frequency, and phase. We compare the proposed uncertainty-aware formulation against (i) an ablated version without uncertainty modeling and (ii) reactive baselines that act only based on the current observations, including Deep Reactive Policy (DRP)~\cite{yang2025deep} and  CuRobo~\cite{sundaralingam2023curobo}. These reactive baselines observe only the current noisy obstacle positions and rely on fast replanning to avoid collisions. We limit the planning time of our method to 66 ms, corresponding to a replanning rate of 15 Hz. For the reactive baselines, we additionally evaluate higher replanning frequencies to examine whether faster replanning can compensate for the lack of future-state prediction.

Table~\ref{tab:dynamic} summarizes the experiment results. Incorporating uncertainty into predictive planning substantially improves robustness to noisy observations and dynamic obstacle motion. Although the uncertainty-aware formulation produces longer trajectories and planning time, it achieves significantly safer behavior in the presence of noisy predictions and dynamic obstacle motion. Increasing the replanning frequency also improves the reactive DRP baseline, but its collision rate remains high even at the highest replanning frequency. CuRobo exhibits lower collision rates at higher replanning frequencies, although it does not successfully complete the task in these highly dynamic scenarios. These results highlight the benefits of predictive control that reasons about future obstacle motion and uncertainty.

\begin{figure*}[hbtp]
    \centering
    \begin{subfigure}[b]{0.16\textwidth}
        \centering
        \includegraphics[trim={6cm 2cm 2cm 2cm},clip,width=1\textwidth]{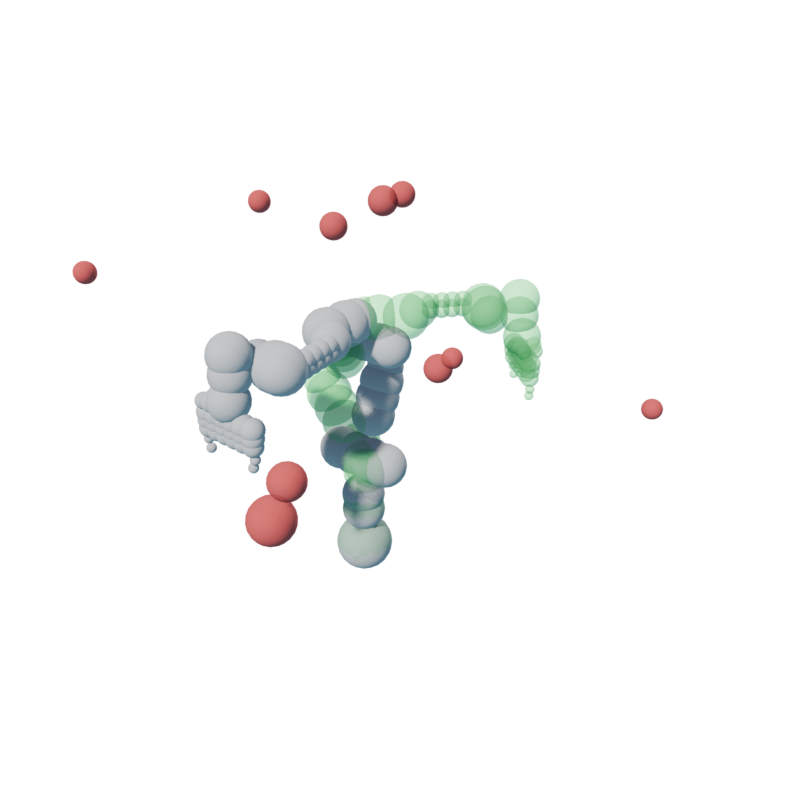}
    \end{subfigure}
    \begin{subfigure}[b]{0.16\textwidth}
        \centering
        \includegraphics[trim={6cm 2cm 2cm 2cm},clip,width=1\textwidth]{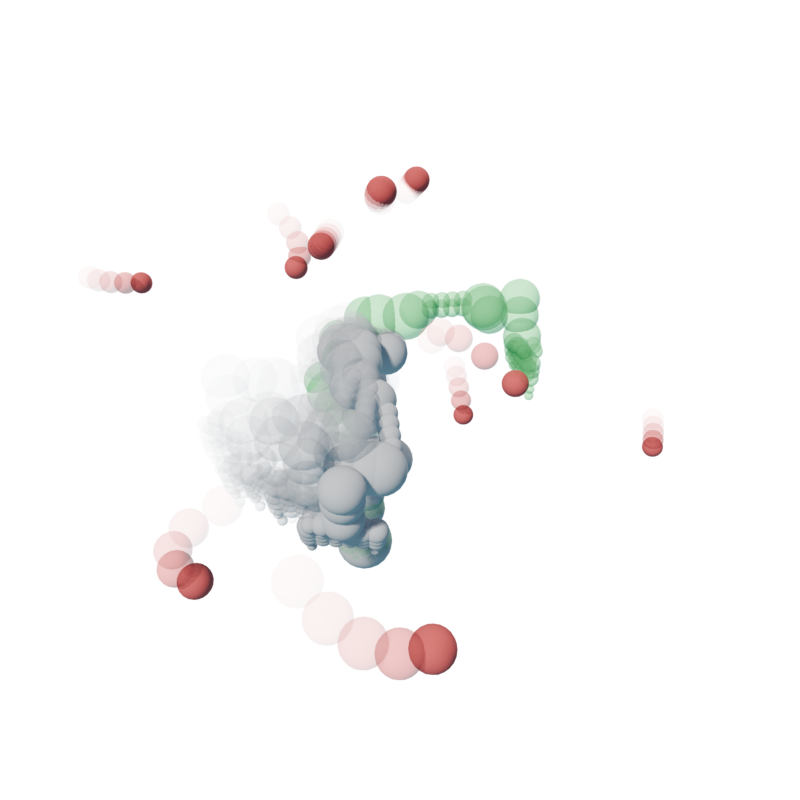}
    \end{subfigure}
    \begin{subfigure}[b]{0.16\textwidth}
        \centering
        \includegraphics[trim={6cm 2cm 2cm 2cm},clip,width=1\textwidth]{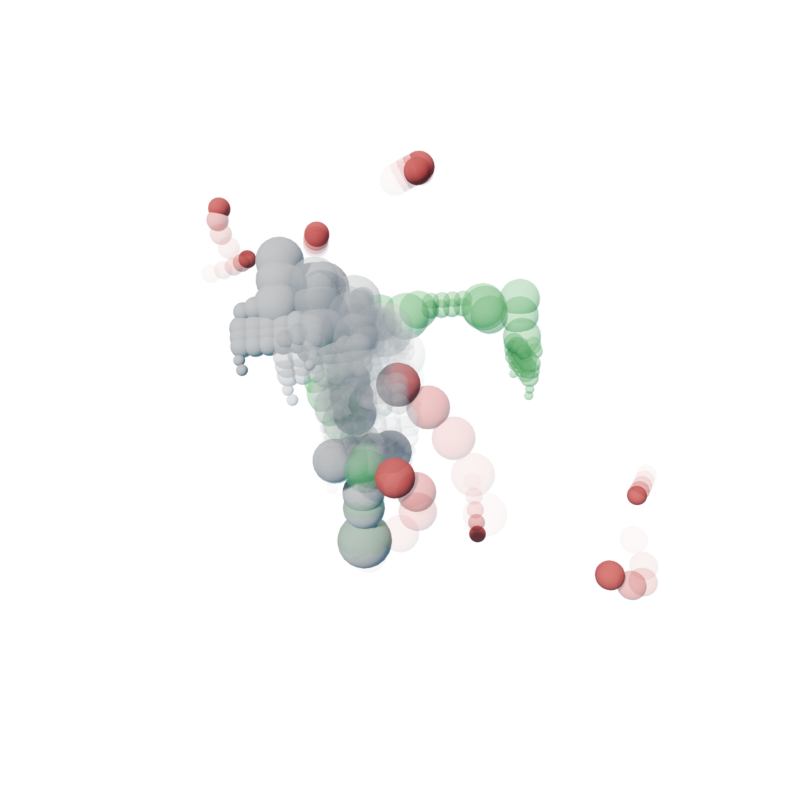}
    \end{subfigure}
    \begin{subfigure}[b]{0.16\textwidth}
        \centering
        \includegraphics[trim={6cm 2cm 2cm 2cm},clip,width=1\textwidth]{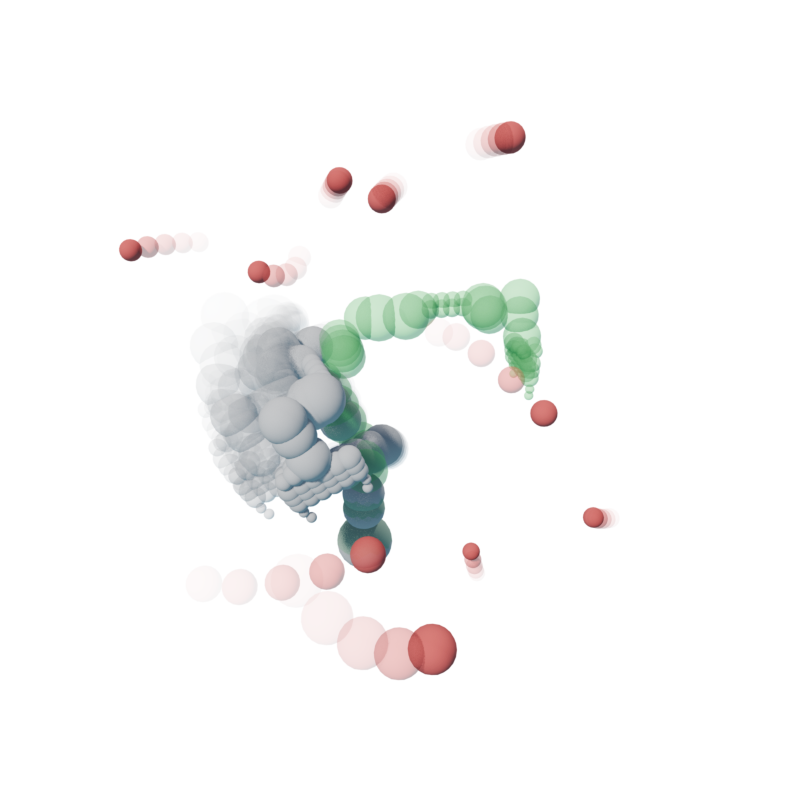}
    \end{subfigure}
    \begin{subfigure}[b]{0.16\textwidth}
        \centering
        \includegraphics[trim={6cm 2cm 2cm 2cm},clip,width=1\textwidth]{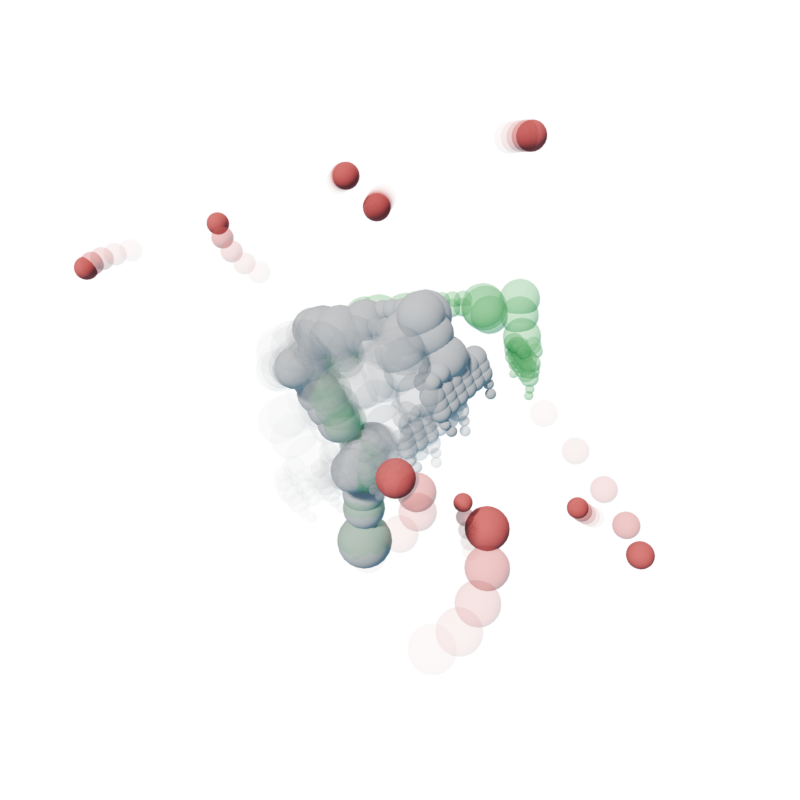}
    \end{subfigure}
    \begin{subfigure}[b]{0.16\textwidth}
        \centering
        \includegraphics[trim={6cm 2cm 2cm 2cm},clip,width=1\textwidth]{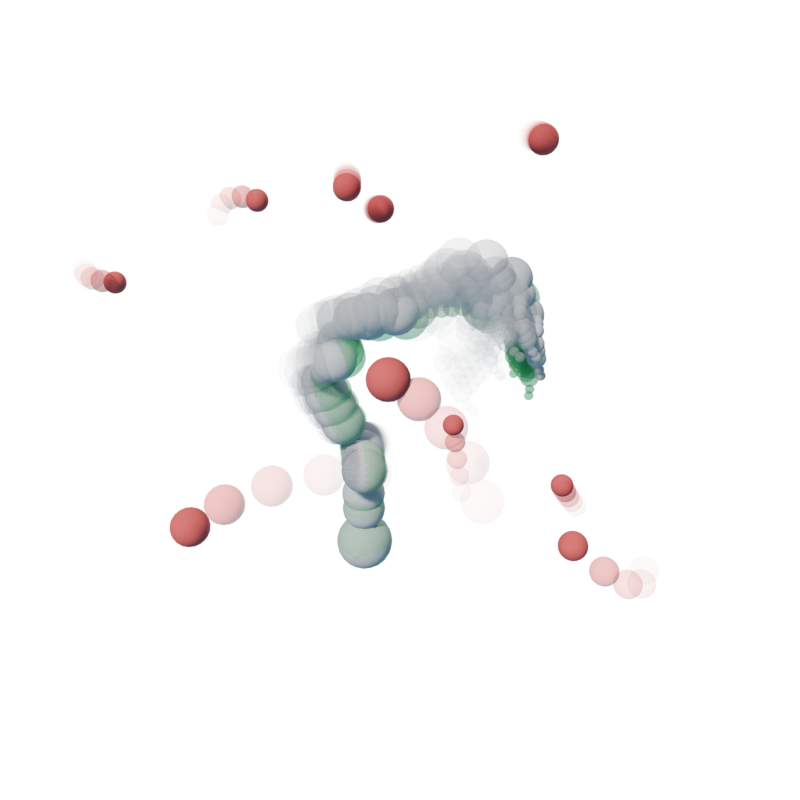}
    \end{subfigure}
    \vspace{-1em}
    \caption{Snapshots of our method successfully planning the robot (solid white) toward the goal (transparent green) among dynamic obstacles (red).  Historical robot and obstacle trajectories are visualized with transparency, with more recent states rendered more opaque.}

    \label{fig:dynamic}
\end{figure*}

\begin{figure*}[hbtp]
    \centering
    \begin{subfigure}[b]{0.19\textwidth}
        \centering
        \includegraphics[trim={0cm 0cm 0cm 4cm},clip, width=1\textwidth]{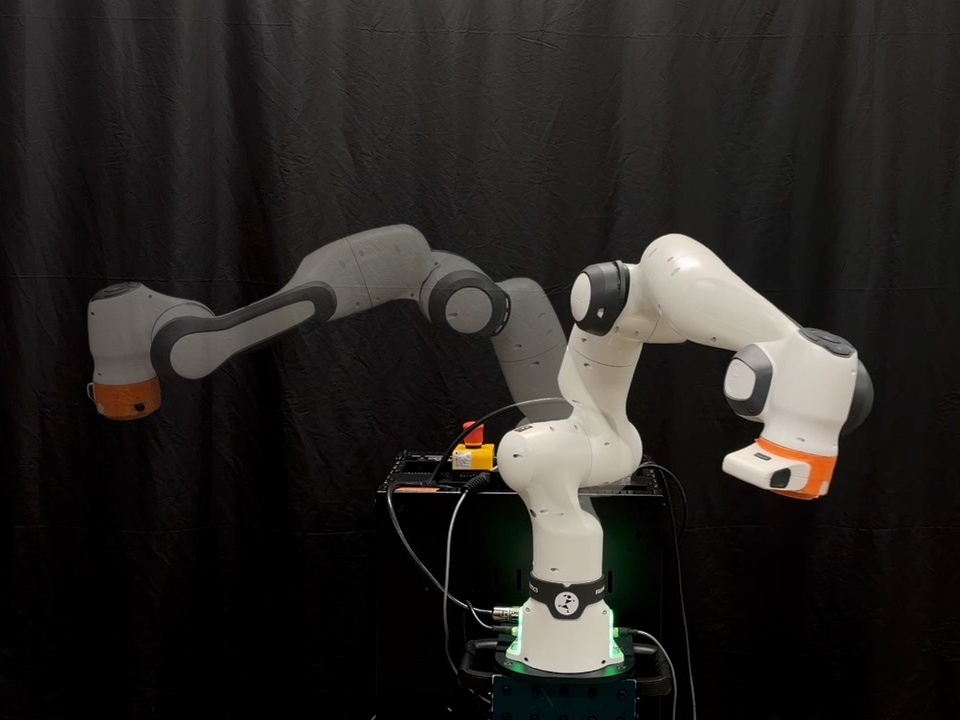}
    \end{subfigure}
    \begin{subfigure}[b]{0.19\textwidth}
        \centering
        \includegraphics[trim={0cm 0cm 0cm 4cm},clip, width=1\textwidth]{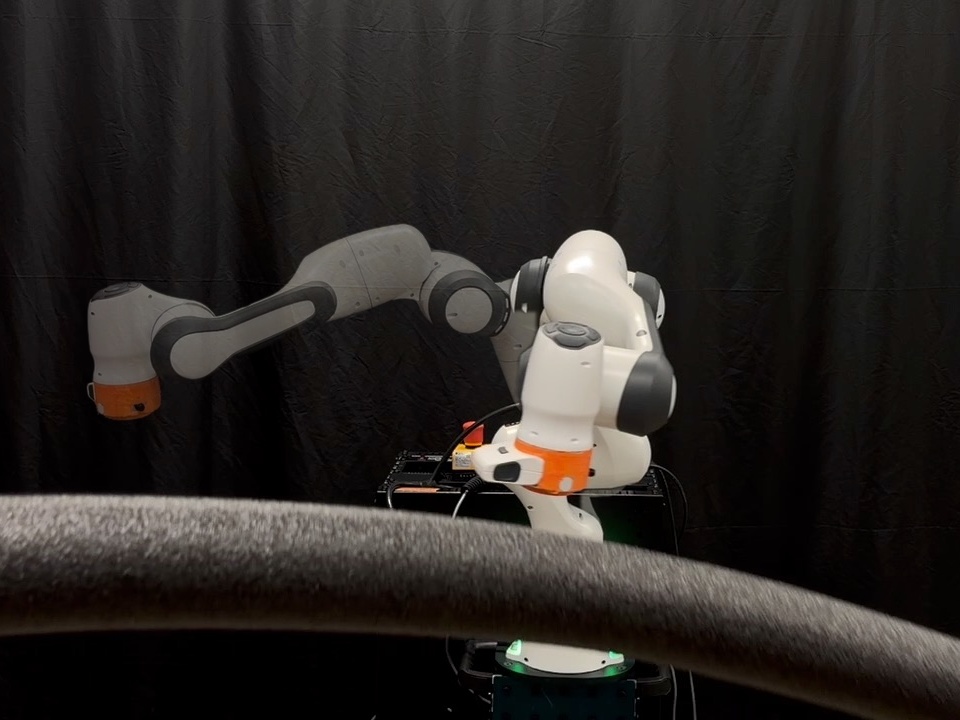}
    \end{subfigure}
    \begin{subfigure}[b]{0.19\textwidth}
        \centering
        \includegraphics[trim={0cm 0cm 0cm 4cm},clip, width=1\textwidth]{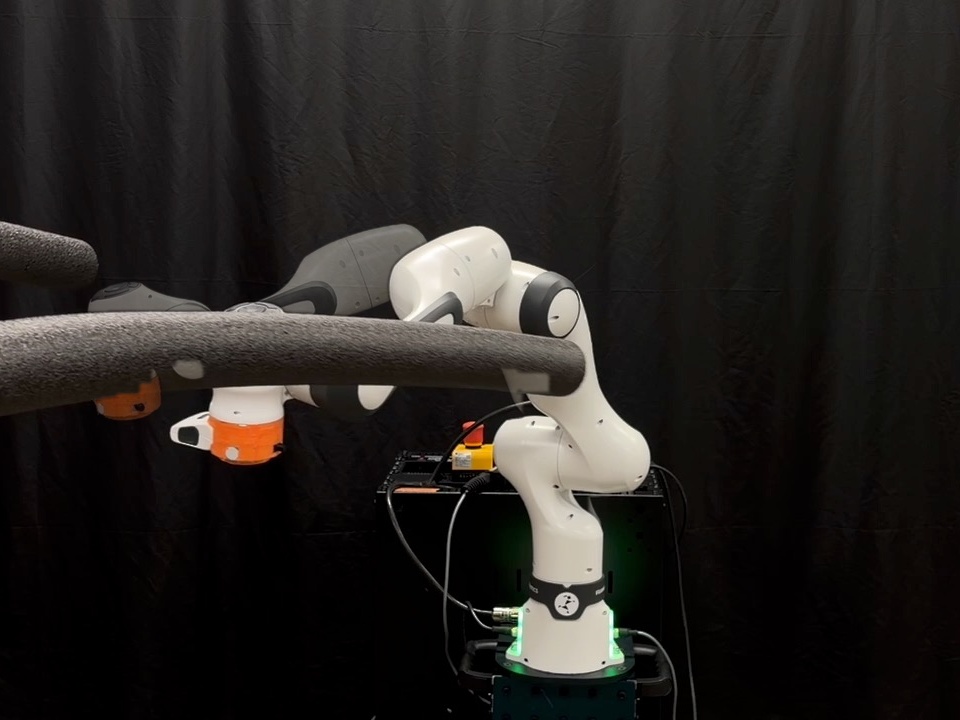}
    \end{subfigure}
    \begin{subfigure}[b]{0.19\textwidth}
        \centering
        \includegraphics[trim={0cm 0cm 0cm 4cm},clip, width=1\textwidth]{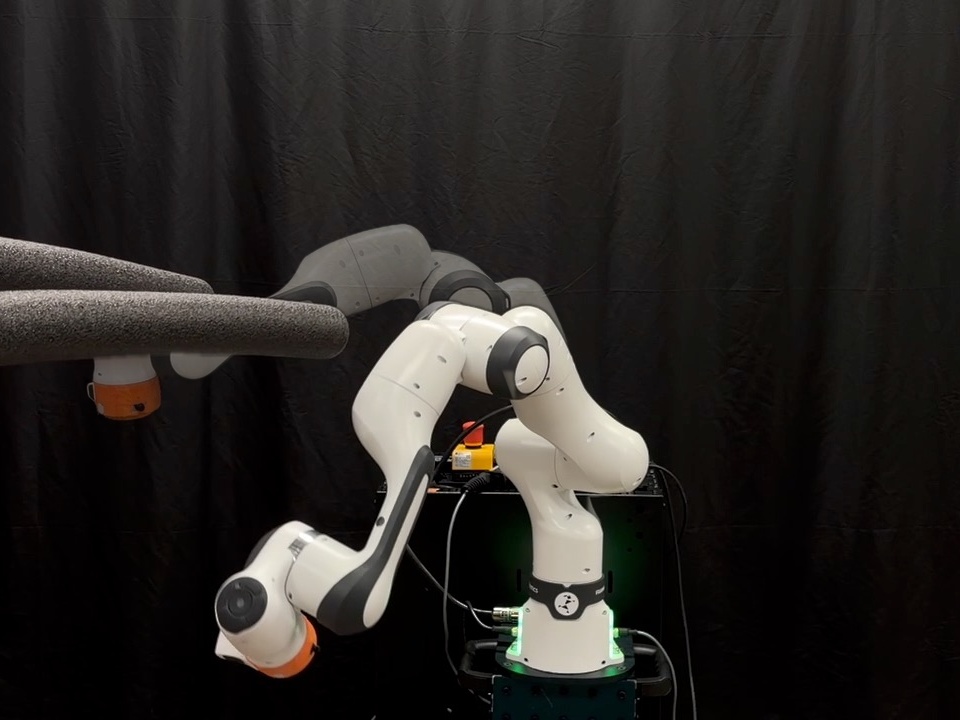}
    \end{subfigure}
    \begin{subfigure}[b]{0.19\textwidth}
        \centering
        \includegraphics[trim={0cm 0cm 0cm 4cm},clip, width=1\textwidth]{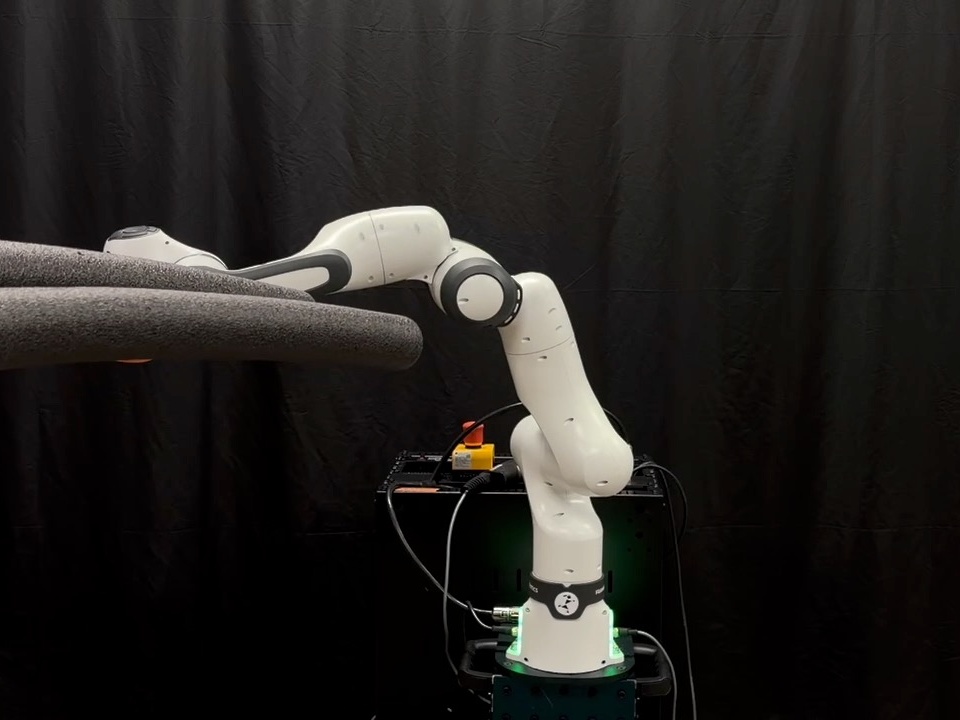}
    \end{subfigure}
    \caption{Real-world dynamic obstacle avoidance on a Franka robot. Snapshots from left to right show the process of the robot avoiding moving foam pool noodles to reach its goal (transparent).}
    \label{fig:hardware}
\end{figure*}

A visualization from the dynamic obstacle avoidance scenario is shown in Figure~\ref{fig:dynamic}. The robot continuously replans under noisy obstacle observations while reasoning about predicted future obstacle motion and uncertainty, successfully reaching the goal while avoiding moving obstacles. To validate the proposed approach under realistic perceptual uncertainty, we further conduct real-world experiments using a Franka FR3 manipulator. The full planner generated velocity commands at approximately 13 Hz with each step taking $78 \pm 19$ ms, which were subsequently executed asynchronously by a configuration level PD controller. Figure~\ref{fig:hardware} shows selected obstacle-dodging behavior. Corresponding videos for both simulated and real-world experiments are provided in the supplementary materials.
\begin{table}[b]
    \centering
    \begin{tabular}{lccccc}
        \toprule
        Method & Hz & SR $\uparrow$ & CR $\downarrow$ & Time [ms] $\downarrow$ & Len $\downarrow$ \\
        \midrule
        \multirow{3}{*}{DRP~\cite{yang2025deep}}
          & 15 & 55 & 44 & \best{9$\pm$4} & 3.3$\pm$0.3 \\
          & 40 & 64 & 36 & \best{9$\pm$4} & 3.7$\pm$0.3 \\
          & 100 & 71 & 29 & \best{9$\pm$3} & 4.3$\pm$0.5 \\
        \midrule
        \multirow{2}{*}{CuRobo~\cite{sundaralingam2023curobo}}
          & 15 & 0 & 24 & 57$\pm$130 & -- \\
          & 40 & 0 & \snd{16} & 56$\pm$91 & -- \\
        \midrule
        SV (MSE) & 15 & 70 & 30 & \snd{28$\pm$14} & \best{2.9$\pm$0.4} \\
        SV (NLL, $\alpha$=0) & 15 & 72 & 28 & 30$\pm$16 & \snd{3.0$\pm$0.6} \\
        SV (NLL, $\alpha$=1) & 15 & \snd{83} & 17 & 40$\pm$21 & 3.4$\pm$0.6 \\
        SV (NLL, $\alpha$=2) & 15 & \best{91} & \best{9} & 46$\pm$21 & 4.4$\pm$1.2 \\
        \bottomrule
    \end{tabular}
    \caption{Dynamic obstacle planning results. Hz: replanning rate, SR: success rate [\%], CR: collision rate [\%], Time: planning time [ms], Len: path length. \best{Best} and \snd{second best} per column.}
    \label{tab:dynamic}
\end{table}

\section{Conclusion}
\label{sec:conclusion}
We present an efficient and uncertainty-aware planning method leveraging probabilistic characterization of a learned swept volume signed distance function. We describe an efficient method to training data using a union of spheres, and propose a framework for estimating and training prediction variance of signed distance functions. Evaluation of the neural swept volume in several cluttered environments shows improvements in planning success while keeping collision rate low. Implementation of the approach on hardware demonstrates its applicability to real-world planning scenarios. 

Despite the empirical performance, several limitations remain and point to future directions. First, the chance-constrained formulation relies on Gaussian assumptions for uncertainty propagation in both the learned model and obstacle perception. In particular, the use of first-order uncertainty propagation introduces an approximation that becomes less accurate in non-linear regimes.
Second, the optimization-based planner does not provide global optimality guarantees. Moreover, because the collision avoidance constraint in ~\eqref{opt:chance-constrained} is incorporated through a soft penalty, constraint satisfaction is not guaranteed and careful weight tuning is required to balance task performance and safety.
Finally, the initial sampling phase in the particle optimization may limit real-time performance with a large number of obstacles. Although it is amenable to acceleration, further work is needed to improve efficiency for resource-constrained robotic systems.

\ifanonymous
\else
\fi

\printbibliography{}

\end{document}